\documentclass[final,5p]{elsarticle}

\usepackage{amsmath,graphicx}

\usepackage[mathcal]{euscript}
\usepackage{amssymb,amsmath}
\usepackage{algpseudocode}
\usepackage[]{algorithm2e}
\usepackage{verbatim}
\usepackage{graphicx}
\usepackage{longtable}
\usepackage{rotating}

\usepackage{bbm}
\usepackage{dsfont}

\usepackage{multirow}
\usepackage{tabularx}
\usepackage{bigints}
\usepackage{tikz}
\usepackage{multirow} 
\usetikzlibrary{arrows,backgrounds,snakes}
\usepackage{amsmath,amstext,amsfonts,amssymb}
\usepackage{amsbsy}
\usepackage{amsthm}
\usepackage{diagbox}
\usepackage{mathabx}
\usepackage{color}
\usepackage{morefloats}
\usepackage{float}
\usetikzlibrary{arrows,shapes,positioning,shadows,trees}
\usetikzlibrary{decorations.pathreplacing}

\usepackage{caption}
\usepackage{subcaption}

\definecolor{darkred}{rgb}{0.5,0.0,0.0}
\definecolor{darkblue}{rgb}{0.2,0.1,0.6}

\newcommand{\ul}{\underline}
\newcommand{\ol}{\overline}

\newcommand{\mc}{\mathcal}
\newcommand{\tc}{\textcolor}

\newtheorem{theorem}{Theorem}[section]

\newtheorem{proposition}[theorem]{Proposition}

\begin{document}

\begin{frontmatter}



\title{Decision-making Oriented Clustering: Application to Pricing and Power Consumption Scheduling}


\author[1]{Chao Zhang\corref{cor1}}
\ead{zhangchaohust@gmail.com}
\author[2]{Samson Lasaulce}
\author[3]{Martin Hennebel}
\author[4]{Lucas Saludjian}
\author[4]{Patrick Panciatici}
\author[5]{H. Vincent Poor}

\cortext[cor1]{Corresponding author}
\address[1]{L2S(CentraleSupelec and Univ. Paris Saclay), 91190 Gif-sur-Yvette, France}
\address[2]{CRAN (CNRS and University of Lorraine), 54000 Nancy, France}
\address[3]{GeePs (CentraleSupelec and Univ. Paris Saclay), 91190 Gif-sur-Yvette, France}
\address[4]{RTE France, 92800 Puteaux, France}
\address[5]{Princeton University, 08544 Princeton, United States}

\begin{abstract}
\tc{black}{Data clustering is an instrumental tool in the area of energy resource management. One problem with conventional clustering is that it does not take the \textbf{final use} of the clustered data into account, which may lead to a very suboptimal use of energy or computational resources. When clustered data are used by a decision-making entity, it turns out that significant gains can be obtained by tailoring the clustering scheme to the final task performed by the \tc{black}{decision-making entity}. The key to having good final performance is to automatically extract the important attributes of the data space that are inherently relevant to the subsequent decision-making entity,  and partition the data space based on these attributes instead of partitioning the data space based on predefined conventional metrics.  For this purpose,  we formulate the framework of decision-making oriented clustering and propose an algorithm providing a decision-based partition of the data space and good representative decisions.  By applying this novel framework and algorithm to a typical problem of real-time pricing  and that of power consumption scheduling, we obtain several insightful analytical results such as the expression of the best representative price profiles for \tc{black}{real-time pricing} and a very significant reduction in terms of required clusters to perform \tc{black}{power consumption scheduling} as shown by our simulations.}
\end{abstract}



\end{frontmatter}
\section{Introduction}
\label{sec:introduction}

It is now well-known that future gains in terms of energy-efficiency will largely rely on the intensive use of data and algorithms. This will be true at a small scale e.g., at the consumer's scale. For instance, charging an electric vehicle (EV) efficiently at home will depend on a forecast of the consumption of the other home appliances. Smart home heating systems will also rely on the exploitation of the data recorded by a smart meter. At a larger scale, transmission operators already have to monitor various energy sources, energy needs of a country and its neighbors, use recorded data to try to predict some key parameters. For all these examples, the number of measurements and even the dimension of the data is typically large and data clustering appears to be an instrumental tool to be able to perform various optimization and decision-making tasks. Clustering is a method that consists in creating clusters, groups, or partitions of data and possibly finding a representative for each cluster. For example, an electricity operator or utility may want to determine e.g., a given number of consumption behaviors and associate a given tariff with each behavior, and for this, properly clustering recorded consumption data is required. 

Above examples are among many others in the area of energy conversion, management, and processing that show the importance of data clustering. Because of its importance, data clustering has become an active research area. Despite of the existence of a quite rich literature, the authors have identified a lack in this area that may make very suboptimal and even non-suitable existing clustering techniques for some key energy management problems such as the power consumption scheduling problem. 

The current conventional data clustering paradigm consists in creating clusters of data based on some similarity indices of various forms. It turns out that the used indices are chosen to be exogenous to the decision-making process that effectively exploits the clustered data, formed clusters, or formed cluster representatives. As a consequence, this may make the decision-making task too complex (e.g., for a human decision-maker), computationally demanding or not admissible, too slow, and very suboptimal. From the physical and technical point of views, adopting the conventional clustering approach may lead to overestimating the amount of required resources e.g., in terms of needed energy, required storage space, transmission bandwidth, or computation capabilities. For instance, an electricity provider or a distribution system operator (DSO) may have no interest in having a very accurate representation of the data recorded by a monitoring device (such as a smart meter). The reason for this may be for designing an implementable pricing policy, for limiting the complexity involved by the optimization tasks at stake (see \cite{Teichgraeber-AE-2019} for a very convincing discussion concerning the interest in using clustering to manage complexity issues in such a scenario), for limiting the amount of private information revealed to the system exploiting the data \cite{sankar-tsg-2012}, or for improving robustness towards diverse forms of noise (see e.g.,\cite{Beaude-TSG-2016} for the problem of forecasting noise). Ultimately, what matters is the quality of the final decision the provider will take (typically from a large number of smart meters).

To bridge the aforementioned gap between conventional clustering and decision-making operations that need to be performed in the area of resource management, we develop a novel framework namely, decision-oriented making clustering (DMOC). To show how the developed framework can be exploited in practice the authors have chosen two important case-studies namely, the power consumption scheduling problem (PCS) and the pricing problem. The PCS problem is a very relevant problem since it appears at different scales of an energy network: at the consumer's scale when an EV or a heating system has to schedule the consumed power so that a need in terms of energy (over a given time window) is filled; at a factory level; at a country level; at a market level when a buyer has to schedule its consumption based on market prices. To be more specific about the limitations of the conventional clustering paradigm, consider the following simplified PCS example. Assume that, after performing some analysis and simplifications due to clustering in particular, the consumer (e.g., an individual or a factory) has the possibility to consume at full power $P$ or not consume at all, and has two periods of time to do this. The power consumption profile can thus only be the sequence $(P,0)$ and $(0,P)$. Measured in terms of similarity (say in terms of Euclidean distance, which would be the case when using the famous k-means clustering technique), these profiles would be declared to be completely different. However, if one assumes that the goal of the decision-maker (the scheduler) wants to minimize the peak-power, these sequences are equivalent. This shows that using a similarity index exogenous to the decision-making process can lead to markedly different outcomes or conclusions. From now on, the authors provide a more detailed view of the existing works and the technical contributions of this article.

In the present paper, the authors' intent is to revisit \tc{black}{the aforementioned dominant data processing paradigm} by assessing the potential benefits from integrating (when available) the knowledge of the final use of the data in the way the data are processed e.g., in the decision-making (DM) process exploiting the data. Because of its importance, the focus of this work is on the problem of data clustering but mathematically, the developed approach is perfectly applicable to signal digitalization and quantization in particular. The most conventional approach of clustering in the area of energy networks is to form groups of data so that the "approximated" data are sufficiently close to the original ones. A typical approach is to make use of similarity or clustering indices such as the Davies-Bouldin index or the Silhouette index (see e.g., \cite{desgraupes-modal-2013} for a review of more than 30 popular indices) to characterize the performance of clustering. However, all of these indices are exogenous to the decision-making process, which effectively exploits the clustered data. \tc{black}{One of the goals we pursue in this paper is to revisit the conventional clustering approach by designing} the clustering scheme so that the cluster or representative information provides sufficient information to perform well in the sense of the performance metric of the decision-maker using the clustered data. \tc{black}{An example of such a decision is to choose in advance a power consumption (PC) profile based on a given \textbf{clustered} day-ahead forecast of the non-controllable part of the PC (NCPC); here, clustering may be applied offline to a database of previous measurements of the NCPC profiles. The problem of electric vehicle charging (at home) would be a typical instance of such scenario: the charging power is controllable whereas, the consumption power associated with the other appliances is assumed to be exogenous to the power control and monitored e.g., through a smart meter.} 

\tc{black}{Below, we review several related works on clustering but it will be seen that none of them adopts the approach of decision-making oriented clustering (DMOC), at least not from the formal point of view developed in this paper. The most famous clustering technique is probably the $k-$means clustering (KMC) technique, which amounts to minimizing a certain Euclidean distance that is clearly independent of how the data are used. KMC has been used e.g., in \cite{Bahl-Energy-2017} for time-series aggregation, in \cite{Ali}{\cite{Li-AE-2021}} to perform load estimation, {in \cite{Tanoto-AE-2020} for electricity generation expansion planning}, or in \cite{Brodrick-Energy-2017}\cite{Gabrielli-AE-2017} where 365 days are clustered into few representative days. More elaborate techniques have been proposed such as Fuzzy C-Means (FCM) clustering to generate the optimal fuzzy rule for decentralized load frequency control \cite{Sudha-Elsevier-2012}, and hierarchical clustering (HC) to aggregate periods with similar load and renewable electricity generation levels \cite{Nahmmacher-energy-2016}\cite{Merrick-EE-2016}{\cite{Tso-AE-2020}}. {To exploit the data features more efficiently, the authors of \cite{Teeraratkul-TSG-2017} proposed to use dynamic time warping instead of the Euclidean distance to partition the residential electricity profiles into different clusters, the authors of \cite{Blanco-TPS-2017} proposed to use cross correlation as a measure to cluster data from wind turbine power generator, {and the authors of \cite{Motlagh-AE-2019} used the delay coordinate embedding technique to reduce the dimensionality of load time series.}} To find appropriate time-series aggregation schemes in energy systems, the authors of \cite{Kotzur-RE-2018} compared the k-means clustering, k-medoids clustering, and hierarchical clustering in presence of an optimization entity. The underlying problem of high time resolution has also been addressed in \cite{pfenninger-AE-2017} and \cite{Teichgraeber-AE-2019}. In \cite{pfenninger-AE-2017}, the focus is on wind and photovoltaic time-series and a planning problem. In \cite{Teichgraeber-AE-2019} the authors consider general complex energy systems in which time-varying operations are performed; they conduct a detailed numerical comparison between conventional clustering (k-means clustering, k-medoids clustering, and hierarchical clustering) and shape-based clustering (dynamic time warping barycenter averaging and k-shape). Notice here, as all aforementioned works, the evaluation is performed ex post, meaning that each given clustering scheme is evaluated in terms of a given objective but not adapted to the objective. {There are also many works on clustering in the computer science literature (even not yet widely applied to energy system problems), but again the existing contributions are data-oriented and not decision-oriented (see e.g., \cite{Zhang-CJC-2002}\cite{Elhamifar-NIPS-2011}\cite{Jain-PRL-2010} \cite{Kriegel-Wiley-2011} \cite{Luxburg-stat-2007}).} The selected references are good representatives of the dominant clustering paradigm, which is either to cluster the data by considering the approximation quality as a primary objective or to cluster to meet imposed constraints (e.g., in terms of complexity).} At last, note that the authors have produced a preliminary work dedicated to a specific quantization problem appearing in wireless communications \cite{zou-wincom-2018} which is partially related to DMOC.  


In contrast with the conventional clustering paradigm, the data attributes are not predefined; the data attributes that are relevant to the decision to be made are automatically extracted by DMOC. To demonstrate the efficiency of the novel approach, the developed framework is applied to two important problems: the problem of real-time pricing (RTP) and the problem of PCS.

The main contributions of this paper can be listed as: (1) we develop a novel data clustering framework in which the partition and representatives are determined under the consideration of the subsequent decision-making operations; (2) we propose the first algorithm to be able to exploit this framework in practice ;(3) we apply this new approach to two important problems in the area of energy networks and provide both analytical and numerical results for these two case-studies; (4) we investigate about the potential improvement the proposed approach can provide when compared to existing state-of-the-art clustering techniques.

The paper proceeds as follows. In Section 2, we introduce the novel framework of DMOC. An alternating optimization algorithm is provided in Section 3 to show how to exploit this framework in practice. In Section 4, the developed framework is applied to two concrete and important problems in the area of energy. Section 5 allows one to assess the potential of DMOC for the two aforementioned problems under a typical simulation setting, and we conclude the article in Section 6.  
 

\textbf{Notation.} Throughout the paper underlined quantities $\ul{v}$, bold quantities $\mathbf{M}$, calligraphic quantities $\mathcal{X}$, and $(.)^{\mathrm{T}}$ will respectively stand for
vectors, matrices, sets, and the transpose operation.
\section{Problem formulation}
\label{sec:pb-formulation}

One considers a database of size $N$. The data set is denoted by $G_N=\{\ul{g}_1,\dots,\ul{g}_N\}$ where $\ul{g}_n \in \mc{G} \subseteq \mathbb{R}^d$ represents Data sample $n \in \mathcal{N}$, $\mathcal{N}= \{1,\dots,N\}$, and $d$ is the dimension of the data space $\mc{G}$. For instance, for the problem of PCS, $\ul{g}_n$ represents the \tc{black}{\tc{black}{NCPC}} profile or vector and $d=T$ is the number of time-slots of the profile (\textcolor{black}{e.g., $T=48$}). Data are clustered in the following sense. The data space is partitioned into cells or clusters denoted by  $\mc{C}_{m}$, $m \in \{1,...,M\}$, $M$ being the number of clusters of the partition. By construction: $\mc{C}_1 \cup \mc{C}_2 \cup \dots \cup \mc{C}_M = \mathcal{G}$ and $\mc{C}_m \cap \mc{C}_{m'} = \emptyset$ for any $m \neq {m'}$. If Data sample $\ul{g}_n$ falls in $\mc{C}_{m}$ then it is represented by the representative $\ul{r}_m \in \mc{R}$, $\mc{R}$ being the space of representatives. For conventional clustering, we typically have that there is a one-to-one mapping between $\mc{R}$ and $\mc{G} $. A key difference between DMOC and conventional clustering is that $\mc{R}$ will correspond to the decision space. A clustering technique or strategy is thus given by a pair under the form $\left\{ (\mc{C}_{m})_{m},  (\ul{r}_{m})_{m}  \right\} $ or equivalently by the clustering operator $\Gamma$ with $\Gamma(\ul{g})  = \ul{r}_m$ when $\ul{g} \in \mc{C}_{m}$, $\ul{g}$ being a generic data sample.

The (most) \textbf{conventional approach} consists in \textcolor{black}{choosing (offline) $\Gamma$ that minimizes the sum of the Euclidean distances between Data sample $\ul{g}_n$ and its representative $\Gamma({\ul{g}_n})$}:
\begin{equation}\label{sec:conv-Q}
\Gamma_{\text{conv}} \in \arg\min_{\Gamma}  \displaystyle\sum_{n=1}^{N}\left\| \Gamma({\ul{g}_n})-\ul{g}_n\right\|^2.
\end{equation}
A way of solving the above minimization problem is to use a (genrally) suboptimal but (generally) implementable technique such as KMC (see e.g., \cite{Ali}). One of the main advantages of such an approach is that it may be possible to obtain explicitly the corresponding partition clusters and the representatives. But this way of clustering data is obviously independent of the final use of the data. For example, if the ultimate goal is to answer a question such as knowing about the absence or presence of a given feature or pattern in the data sample, partitioning the data space in two clusters only may be sufficient and, the way to split the space has to be made according to the considered final feature detection performance metric. More generally, if the task performed by the DM entity is known, it seems to be possible to improve the clustering technique (e.g., by decreasing the number of clusters or by improving its approximation quality). This is precisely the approach adopted in this paper.

Formally, the \textbf{proposed approach} (see Fig.~1) consists in assuming that the (online) task to be performed by the DM entity (e.g., a power consumption scheduler) can be represented \textcolor{black}{by a standard OP, that is, a given function has to be maximized under some constraints}. Therefore, the goal is to maximize a certain function or performance metric $f(\ul{x};\ul{g})$ (e.g., some profit or revenue function) with respect to the decision variable $\ul{x} \in \mathcal{X}$, $\mathcal{X} \subseteq \mathbb{R}^D$, $D\geq 1$, given some measurement of the parameters $\ul{g}$ under some constraints under the form $d_i(\ul{x}) \leq 0$, $i\in\mathcal{I}$, $\mathcal{I}=\{1,...,I\}$, and $e_j(\ul{x}) = 0$, $j\in\mathcal{J}$, $\mathcal{J}=\{1,...,J\}$. This mathematically writes as the following standard form \textbf{online OP}:
\begin{equation}
\begin{split}
\underset{\ul{x}}{\mathrm{minimize}}&\quad -f\left(\ul{x} ; \ul{g}\right)\\
\mathrm{s.t.} & \quad d_i(\ul{x})\leq 0, i\in\mathcal{I}\\
&\quad e_j(\ul{x})=0, j\in\mathcal{J}
\end{split}.
\label{eq:sec2_1}
\end{equation}
By denoting $\ul{x}^{\star}(\ul{g})$ an optimal solution of the above OP, the problem of finding a DMOC scheme therefore amounts to solving the following \textbf{offline problem}:
\begin{equation}\label{sec:new-Q}
\Gamma_{\text{new}} \in \arg\max_{\Gamma}  \displaystyle\sum_{n=1}^{N} f(\ul{x}^{\star}( \Gamma(\ul{g}_n) )  ; \ul{g}_n).
\end{equation}

\tc{black}{At this point, the difference between the conventional clustering paradigm and the DMOC paradigm appears very clearly:}
\begin{itemize}
\item \tc{black}{The conventional clustering paradigm: 1. exploits, in an offline manner, the data set $\mc{G}_N$ to compute the partition of the \textbf{data} space $\mc{G}$ and the representative data points (e.g., with KMC); 2. uses, in an online manner, clustering to find the representative $\widehat{\ul{g}}$ of the current data sample $\ul{g}$; 3. solves, based on the knowledge of $\widehat{\ul{g}}$, the OP which determines the best decision $\ul{x}$ (namely, maximizing $f$ under some constraints).}
\item \tc{black}{The DMOC paradigm: 1. exploits, in an offline manner, the data set $\mc{G}_N$ to compute (via solving OP (\ref{eq:rep_ori})) the partition of the \textbf{decision} space $\mc{X}$ and the representative decision points; 2. uses, in an online manner, DMOC to directly find the representative (final) decision $\widehat{\ul{x}}$ associated with the current data sample $\ul{g}$.}
\end{itemize}
It can be checked that the conventional clustering approach given by (\ref{sec:conv-Q}) can be obtained from (\ref{sec:new-Q}) by making the following specific choices: $\mathcal{R}=\mathcal{G}$; $f(\ul{x};\ul{g})=-\|\ul{x}-\ul{g}\|^2$, $d_i(\ul{x}) = -\infty$, and $e_j(\ul{x}) = 0$ for all $(i,j) \in \mathcal{I} \times \mathcal{J}$. In its full generality, solving the problem associated with (\ref{sec:new-Q}) is difficult. Indeed, finding the best clusters and the best representatives jointly may be hard both from the analytical and computational point of view. This is one of our motivations for proposing an alternating optimization algorithm in the next section. 

\begin{figure}[h]
   \begin{center}
        \includegraphics[width=.46\textwidth]{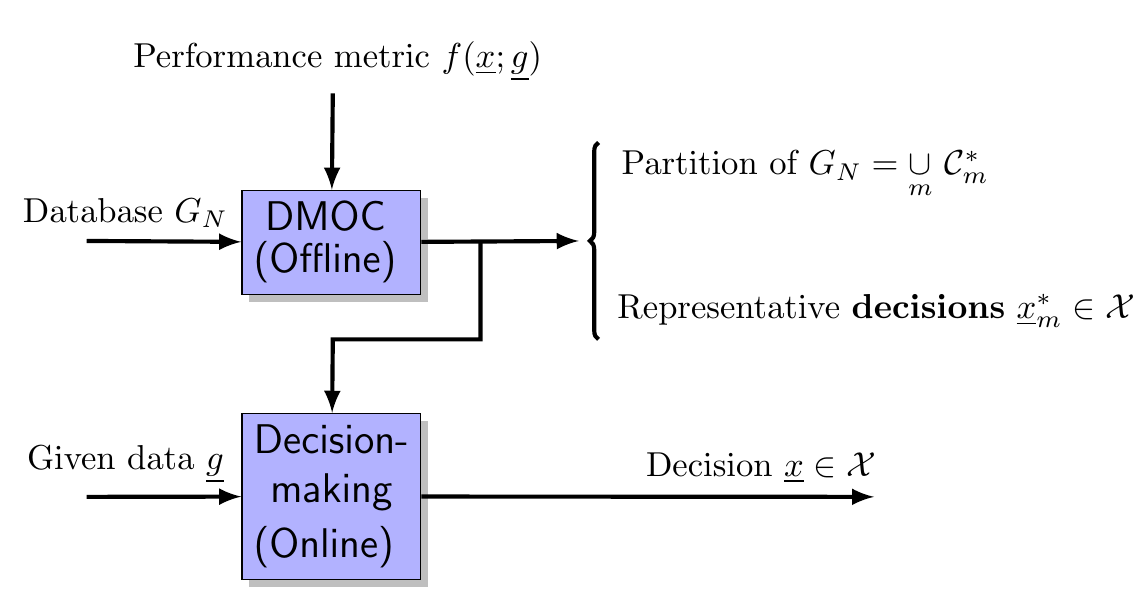}
    \end{center}
    \caption{The DMOC approach. As key points notice that: 1. An \textbf{arbitrary} DM performance metric can be considered ($f$); 2. The representatives of the clusters are in the \textbf{decision} space $\mathcal{X}$; 3. What matters for a DM entity aiming at maximizing $f(\ul{x};\ul{g})$ w.r.t. the decision $\ul{x}$ given $\widehat{\ul{g}}$ (clustered/imperfect/noisy version of $\ul{g}$ is not the similarity between $\ul{g}$ and $\widehat{\ul{g}}$ but the average of the \textbf{final optimality loss} measured by $| f(\ul{x}^\star(\ul{g});\ul{g}) - f(\ul{x}^\star(\widehat{\ul{g}});\ul{g}) |$.}
\end{figure}

\tc{black}{\textbf{Remark 1}. In this paper, we have selected as case studies the problem of RTP and PCS. For RTP, the decision-maker is the provider, its decision consists in choosing electricity price profiles or tariffs, and the measured data consists of the various satisfaction parameters of the consumers. For PCS, the decision-maker is a consumer (e.g., a factory or an electric vehicle), its decision consists in choosing a power consumption profile, and the measured data consists of the non-controllable power consumption profiles. Many other problems of the modern power grid are concerned by the newly developed framework and would need to be studied from the proposed perspective. For example, electricity market price profiles may be clustered for a given purpose and the (distribution/transmission) network states may be clustered to be able to characterize its behavior (e.g., the absence/presence of a global anomaly).}

\section{Proposed algorithm}
\label{sec:proposed-algorithm}

As mentioned in the previous section, solving the OP associated with (\ref{sec:new-Q}) is not an easy task in general. In fact, even for a specific performance metric $f(\ul{x};\ul{g}) = -\|\ul{x}-\ul{g}\|^2$ which is used for $k-$mean like algorithms, it is known that some degree of suboptimality has to be accepted. In the present section, we propose an algorithm which applies to any performance metric $f$ and relies on two key ingredients. First, by providing an appropriate equivalence argument, the problem of finding the representatives of the data is converted into a problem of finding the representative \textbf{decision} points. Second, since the joint determination of the optimal clusters and representative decision points is difficult in general, we resort to an iterative (and suboptimal) algorithm which operates in two steps. 

When inspecting (\ref{sec:new-Q}), it is seen that the optimal DM function $\ul{x}^{\star}(\ul{g})$ is needed. Although there are well-known examples for which such a function can be found (e.g., the valley-filling solution \cite{gan-2012}), this knowledge is not always available. This is one of the reasons why we reformulate the problem of finding the data representatives $\ul{r}_1,...,\ul{r}_M$ into that of finding representative decision points $\ul{x}_1,...,\ul{x}_M$. The equivalence between these two problems is the purpose of the proposition below. Before stating this proposition, a few notations are in order. The set of data indices is assumed to be partitioned as follows: $\mathcal{N}=\mathcal{N}_1\cup\mathcal{N}_2\cup\dots\cup\mathcal{N}_M$. For $m\in\{1,...,M\}$, the set $\mathcal{N}_m$ represents the set of indices of the data samples which belongs to Cluster $\mathcal{C}_m$, \textcolor{black}{i.e., $\ul{g}_n\in\mathcal{C}_m$ implies $n\in\mathcal{N}_m$ and vice versa}. Therefore, the set $\mathcal{N}_m$ completely characterizes the cluster $\mathcal{C}_m$ and conversely. For $\ul{g} \in \mathcal{C}_m$, the representative decision point is denoted by $\ul{x}_m$. Using these notations, the following result can be stated. 

\begin{proposition}
The offline OP associated with (\ref{sec:new-Q}) is equivalent to the following offline OP:
\begin{equation}
\begin{split}
\underset{\left( \mathcal{N}_1,..., \mathcal{N}_M, \ul{x}_1,..., \ul{x}_M  \right)  }{\mathrm{minimize}}&\quad-\sum_{m=1}^M\sum_{n\in \mathcal{N}_m} f(\ul{x}_m; \ul{g}_n)\\
\mathrm{s.t.}&\quad d_i(\ul{x}_m)\leq 0, i \in \mathcal{I}\\
&\quad e_j(\ul{x}_m)=0, j \in \mathcal{J}
\end{split}
\label{eq:criterion}
\end{equation}
\label{prop:1}
\end{proposition}
\begin{proof}
See Appendix.
\end{proof}
Prop.~III.~1 allows one to characterize the optimal DMOC strategies. But, for classical complexity arguments, we resort to an alternating optimization algorithm to find $(\mathcal{N}_1, ..., \mathcal{N}_M)$ and $(\ul{x}_1, ...,\ul{x}_M)$ in an iterative manner. For this, it is first assumed that a set of representative decision points is given. It can then be checked that the best clusters (given a set of representative decision points) are given by:
\begin{equation}
\mathcal{C}_m^*=\left\{ {g_n} \in \mathcal{G}  : \,f(\ul{x}_m;{\ul{g}_n})\geq f(\ul{x}_{m'};{\ul{g}_n}) \,\,\,\forall m',\,\,\, 1\leq m'\leq M\right\}.
\label{eq:vq2_vector}
\end{equation}
\textcolor{black}{or equivalently,} 
\begin{equation}
\mathcal{N}_m^*=\left\{ n \in \mathcal{N}  : \,f(\ul{x}_m;{\ul{g}_n})\geq f(\ul{x}_{m'};{\ul{g}_n}) \,\,\,\forall m',\,\,\, 1\leq m'\leq M\right\}.
\label{eq:vq_vector_n}
\end{equation}
For the sake of clarity, we will mainly use the notation $\mathcal{C}_m$ to refer to Cluster $m$. The above formula characterizes the optimal clusters for given representative decision points. To know more about the "geometry" of the clusters, a specific choice for $f$ has to be made. For instance, for $f(\ul{x};\ul{g})=-\|\ul{x}-\ul{g}\|^2$ and $N$ large, the best clusters correspond to the famous Vorono\"{i} regions. Now, as a second step, we now assume that some choice for the clusters is made and want to characterize the representative decision points which maximize the considered performance metric. It can be checked that for $m\in\{1,...,M\}$, the best representative decision (given a set of clusters) is obtained by solving the following OP:\begin{equation}
\begin{split}
\ul{x}_m^*\in&\quad -\underset{\ul{x}\in\mathcal{X}}{\arg\min}\displaystyle\sum_{n\in {\mathcal{N}_m}} f(\ul{x}_m;\ul{g}_n)\\
\mathrm{s.t.}&\quad d_i(\ul{x}_m)\leq 0, i\in \mathcal{I}\\
&\quad e_j(\ul{x}_m)=0, j\in \mathcal{J}.
\label{eq:rep_ori}
\end{split}
\end{equation}

Equations (\ref{eq:vq2_vector}) and (\ref{eq:rep_ori}) precisely constitute the two steps of Algorithm 1, which is the iterative algorithm proposed to determine the clusters and representative decision points to solve the original OP given by (\ref{sec:new-Q}). These two steps are performed at each iteration of the algorithm until convergence is reached. At each iteration, the function to be maximized in (\ref{eq:rep_ori}) can only increase or stay constant. Since functions of practical interest are generally bounded, convergence is guaranteed. Similarly to iterative algorithms such as KMC, convergence to a global minimum is not guaranteed in general. \tc{black}{Maximizing jointly a function w.r.t. the set of clusters and the set of representatives is known to be an NP-hard problem (See \cite{Garey-TIT-1982}\cite{Hanna-JSAIT-2020}). This is the reason why we resort to an alternating optimization algorithm. In general, the proposed DMOC algorithm guarantees convergence to a local maximum. In the scalar case (namely, $g \in \mathbb{R}$ as it is the case for the RTP case) sufficient conditions under which convergence to a global maximum may be exhibited. For instance, this is the case when $f(x;g)=-(x-g)^2$ and the probability distribution function $\phi(g)$ is log-concave \cite{Fleischer-TIT-1964}. In particular, if $g$ is normally distributed, global convergence is available. Another interesting case is when the function verifies the following property $f(x;g)=f(x-g;g)$. Then, reference \cite{TRUSHKIN-TIT-1982} allows one to claim that the maximum point is unique, which guarantees global convergence. In the vector case (as in the PCS case), the problem becomes more complicated. In particular, finding a general way to determine the optimal tesselation structure of the clusters is known  to be an open problem \cite{Gray-TIT-1998}. Notice that the proposed iterative algorithm can always be initialized with the best state-of-the-art solution. This guarantees a positive performance gain over any state-of-art solution. This value of this positive gain will be assessed thanks to the detailed numerical performance analysis conducted in Sec.~\ref{sec:num-perf-analysis}. To conclude on the proposed algorithm, note that} to run the algorithm, only the data set $G_N$ and a given initial choice of the representative decision points are required. The maximum number of iterations $Q$ is fixed. 

\begin{algorithm}[h]
{\bf{Inputs:}} Data set $G_N$; initial representative decisions $\{\ul{x}_1^{(0)},...,\ul{x}_M^{(0)}\}$; number of clusters $M$; number of iterations $Q$\\
{\bf{Outputs:}} $\{\ul{x}_1^{*},...,\ul{x}_M^{*}\}$, $\{\mc{C}_1^{*},...,\mc{C}_{M}^{*}\}$\\ 
{\bf{Initialization:}} Set iteration index $q=0$. Initialize the representatives by $\{\ul{x}_1^{(0)},...,\ul{x}_M^{(0)}\}$. {Set performance evaluation quantity $A_0=0$ and $A_{-1}=-100$. Set the tolerance as $T_d$}. \\
\While{$q<Q$ and $A_q-A_{q-1}>T_d$}{
Update the iteration index: $q \gets q+1$.\\
For all $m \in \{1,...,M\}$, update
$\mc{C}_m^{(q)}$ from $\ul{x}_m^{(q-1)}$ using (\ref{eq:vq2_vector}).\\
For all $m \in \{1,..,M\}$, update $\ul{x}_m^{(q+1)}$ for each cluster $\mc{C}_m^{(q)}$ by solving OP (\ref{eq:rep_ori}).\\
Compute $A_q=\displaystyle\sum_{n=1}^N\sum_{m=1}^M f(\ul{x}_m^{(q)};\ul{g}_n)\mathbbm{1}_{\ul{g}_n\in\mc{C}_m^{(q)}}$.
}
$\forall m \in \{1,...,M\}, \,\,\, \ul{x}_m^{*}= \ul{x}_m^{(q)}$, $\mc{C}_{m}^{*}= \mc{C}_{m}^{(q)}$
\caption{\small Algorithm to obtain a DMOC strategy} \label{algo_vector}
\end{algorithm}

\begin{figure}[h]
   \begin{center}
        \includegraphics[width=.48\textwidth]{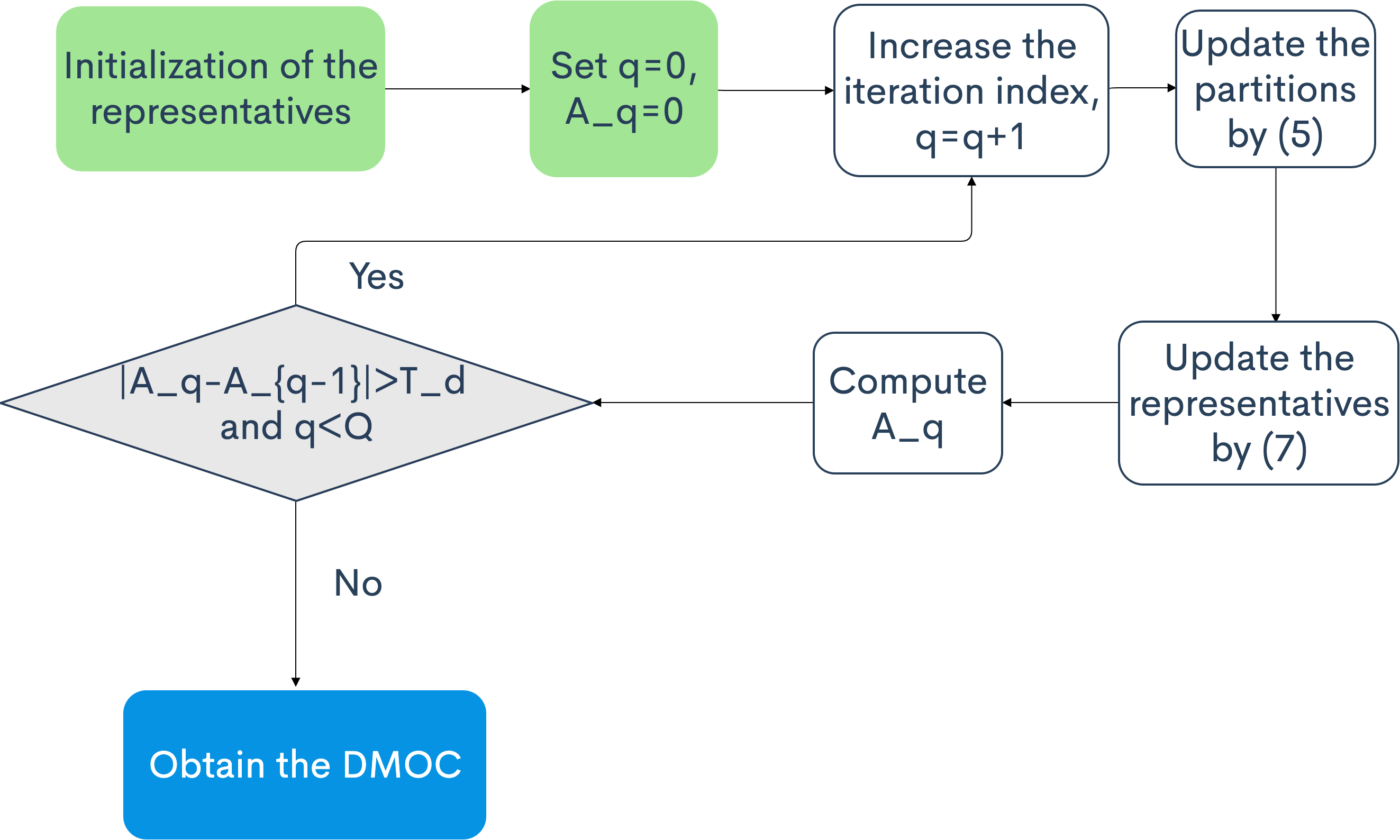}
    \end{center}
    \caption{Flowchart of the proposed algorithm}
\end{figure}

In the next section, we show how to exploit this general algorithm for specific performance metrics. 

\tc{black}{\textbf{Remark 2}. We will not conduct here the complexity analysis of Algorithm 1, the main objectives of this paper being to introduce the DMOC approach, to provide one possible algorithm to implement it, and to assess the performance gains (measured in terms of a given utility function) for the two applications of interest. Nonetheless, we would like to make some useful comments on this issue. As illustrated through Fig.~1, offline operations have to be distinguished from online operations. Algorithm 1 only relies on offline operations namely, computing the partition $\{\mc{C}_m^*\}_{m=1}^M$ and the representative decisions $\{\ul{x}_m^*\}_{m=1}^M$. This means that these operations can be made once and for all by powerful computers. In fact, even for the most computationally demanding scenario studied in Sec. V, the total computation time has never exceeded 10 min with a standard personal computer. As for the online operation, given $\ul{g}$, it consists in selecting the element in the set $\{\ul{x}_m^*\}_{m=1}^M$ that maximizes $f(\ul{x}_m^*;\ul{g})$. The complexity of this operation is in $O(M)$.}

\section{Applying DMOC to Real-time Pricing and Power Consumption Scheduling}
\label{sec:case-studies}

In this section, we make specific choices for the performance metric $f(\ul{x};\ul{g})$. We have selected the two considered corresponding metrics because they concern quite a large number of works in the literature of smart grids and also because they allow us to clearly illustrate the new point of view developed in this paper. The first metric corresponds to a largely used performance criterion which is derived from \cite{MR-SGC-2010}. It consists in mixing the social welfare of a group of consumers and the total production cost; other relevant pricing problems (e.g., \cite{yang-tsg-2018}) might be considered as extensions of this work. The decision for the provider corresponds to a price profile, pricing strategy, or tariff policy and the function parameters to the satisfaction parameters for the consumers. The second performance metric corresponds to the Lp norm, which in particular allows one to include as a special case the peak power minimization problem. Here, the decision is a power consumption profile and the parameters correspond to the non-flexible part of the power consumption.

\subsection{Pricing problems}

\tc{black}{The first step taken in this section is to study pricing problems with clustering. Since operating at high time resolution over a long period of time generally leads to intractable optimization problems \cite{Teichgraeber-AE-2019,Kotzur-RE-2018, pfenninger-AE-2017}, we resort to clustering. Clustering is applied here to obtain a small number of representative time profiles (e.g., cluster the 365 days into 5 representative time profiles) and to design a corresponding tariff for each representative. We derive the performance metric given by (\ref{eq:welfare}) from the largely used RTP setting proposed in \cite{MR-SGC-2010}. For this model, we are able to calculate the optimal tariff $\ul{x}$ from consumers' loads/demands time-series $\ul{g}$. We consider a provider and a set of consumers $\mathcal{K}=\{1,\dots,K\}$. Our goal is to cluster the corresponding time-series and associate with each cluster a representative tariff, the association being performed by using the DMOC algorithm of Sec. III. The considered performance metric for the provider implements a tradeoff between the sum of the consumer's utility functions and the total energy procurement cost (\ref{eq:welfare}). The action of the provider at time $t$ is denoted by $x(t) \geq 0$ and consists in choosing the price of electricity at time $t \in \mathcal{T}$, $\mathcal{T}=\{1,...,T \}$, $T$ being the number of time-slots of the time period under consideration (\tc{black}{For instance, a time period can consiste of a day and a time-slot can consist of an hour if $T=24$})}. \tc{black}{Time-varying prices are natural when the consumers correspond to large entities such as states or big companies but are also well motivated for future grids when they correspond to individuals (see e.g., \cite{MR-SGC-2010}\cite{Hao-TPS-2012}).}. To take his decision, the provider has some knowledge about the satisfaction parameters of the consumers (see \cite{MR-SGC-2010} for more details) at the current time-slot. The satisfaction parameter for Consumer $k \in \mathcal{K}$ at time $t \in \mathcal{T} $ is denoted by $g_k(t) \geq 0 $. It is defined through the generic benefit or utility function $u$ for the consumers:
\begin{equation}\label{eq:ee}
u(\ell ; g)=
\left|
\begin{array}{ll}
g \ell-\frac{\alpha}{2} \ell^2 & \hbox{  if \,\,\,$0\leq \ell \leq \frac{g}{\alpha}$} \\\\
\frac{g^2}{2\alpha} & \hbox{ if \,\,\,$\ell> \frac{g}{\alpha}$} \\\\
\end{array}
\right.
\end{equation}
where $\ell\geq 0 $ represents the load or consumption level, $g$ the satisfaction parameter, and $\alpha >0 $ is a constant.  This means that the benefit of the consumer increases quadratically with the load level but reaches a saturation point determined by the parameters $g$ and $\alpha$. To make this generic utility consumer-specific and time-dependent, one just has to replace $g$ with $g_k(t)$ and $\ell$ with $\ell_k(t)$. As in \cite{MR-SGC-2010}, it is assumed that Consumer $k$ has a cost for consuming under the form $x(t) \ell_k(t) $ where $x(t)$ represents the price of electricity at a given time $t$. The generic combined utility for the consumer thus writes as the difference $u(\ell_k(t) ; g_k(t)) - x(t) \ell_k(t)$. For a given price of electricity chosen by the provider at time $t$, the combined utility is assumed to determine the best response in terms of consumption, which is to consume at a load level given by:
\begin{equation}\label{eq:power_op2}
\ell^{\star}_k(x(t))=
\left|
\begin{array}{lll}
0& \hbox{  if \,\,\,$x(t)>g_k(t)$} \\\\
\frac{g_k(t) -x(t) }{\alpha} & \hbox{ if \,\,\,$x(t)\leq g_k(t)$} \\\\
\end{array}.
\right.
\end{equation}
Of course, when the price is higher than the satisfaction parameter, $\ell^{\star}_k$ is not an interior solution and reaches the minimal consumption level allowed and therefore has to be replaced with the corresponding value. For the sake of clarity, we assume no over-pricing for the case study under consideration, i.e., $\ell^{\star}_k(x(t))=\frac{g_k(t) -x(t) }{\alpha}$. To design its DMOC strategy, the provider is assumed to pursue (possibly by using a learning algorithm which only exploits partial or indirect information e.g., about $g_k(t)$) the maximization of the average welfare of all its customers over time minus the cost of energy procurement (quadratic cost model) as follows~: 
{\small \begin{equation}
f_{1}(\ul{x}; \ul{g}) =\sum_{t=1}^{T} \left[ \sum_{k=1}^{K} u(\ell^{\star}_k(x(t)) ; g_k(t) ) - a  (L^{\star}(t))^2   - b L^{\star}(t) - c \right]
\label{eq:welfare}
\end{equation}}
where $\ul{x} = (x(1),\dots,x(T))$ (i.e., $d=T$) is the sequence of prices chosen by the provider;
\begin{equation}
\ul{g} = (g_1(1),\dots,g_K(1), \dots, g_1(T),\dots,g_K(T) )
\end{equation} is a vector which comprises all the consumer satisfaction parameters in one time period; $L^{\star}(t) = \sum_{k=1}^K \ell_k^{\star}(t)$ is the total load induced by the $K$ consumers for time-slot $t$ and $(a,b,c)$ is a triplet of constants to model the (quadratic) procurement cost for the provider. {The constraint on RTP is the positivity of price, namely, \begin{equation}
x(d)>0,\quad \forall d\in\{1,\dots,D\}.
\end{equation}}\tc{black}{Our target is to cluster the set $\mathcal{G}=\{\ul{g}_1 ,\dots, \ul{g}_N \}$ consisting of $N$ time period vectors with
\begin{equation}
\ul{g}_n=(g_1^{(n)}(1),\dots,g_K^{(n)}(1), \dots, g_1^{(n)}(T),\dots,g_K^{(n)}(T) )
\end{equation}
into $M$ representative time periods, and find the corresponding tariff for each representative time period.} For the performance metric $f_1$, it turns out that the clusters and representatives respectively obtained by the general equations (\ref{eq:vq2_vector}) and (\ref{eq:rep_ori}) express in an elegant manner. \tc{black}{Before providing the corresponding proposition, let us introduce some auxiliary quantities. Notice that the parameters $K$, $\alpha$, $a$, $b$, and $c$ have all been defined in the current subsection. We introduce three scalar quantities: $\widetilde{a} ={\frac{K^2}{\alpha^2} \left( a+ \frac{\alpha}{2K} \right)}$; $\kappa = {\frac{aK}{\alpha^2 \widetilde{a} }}$; $\beta = {\frac{bK}{2\alpha \widetilde{a} }}$. From this and by denoting  $\underline{1}_K$ the column vector of $K$ ones, we define the two following quantities: $\underline{\beta}= \beta \underline{1}_T$; $\mathbf{A} = \kappa \mathbf{I}_T \otimes \underline{1}_K^{\mathrm{T}}$, the operator $\otimes$ standing for the Kronecker tensor product \cite{Matrix-cookbook}.} 

\begin{proposition}[DMOC for RTP] For a given sequence of representative price profiles $\underline{x}_{m} $ the best way of clustering (in the sense of $f_1$) the consumer's satisfaction parameter space is given by the following clusters: 
\begin{equation}
\begin{array}{ll}
\tc{black}{\mc{C}_m^*= \left\{ {\underline{g}} \in \mathcal{G}~: \,\| \mathbf{A} \underline{g}  + \textcolor{black}{\underline{\beta}}    -   \underline{x}_m  \|_2^2\leq  \| \mathbf{A} \underline{g}  + \textcolor{black}{\underline{\beta}}     -   \underline{x}_{m'}  \|_2^2, \right.}\\
\tc{black}{ \left. \forall m'\neq m \right\} }. 
\end{array}
\label{eq:cell_caseII}
\end{equation}
Now, for a given partition of the satisfaction parameter space into a set of clusters $\mc{C}_m$, the best representative price profiles are given by:
\begin{equation}
x_m^*(t) = \frac{a \ol{g}_m(t) + \frac{\alpha b}{2K} }{a + \frac{\alpha}{2K}}
\label{eq:best_representative_case1}
\end{equation}
where $\displaystyle{\ol{g}_m(t) = \frac{ \sum_{n=1}^N\sum_{k=1}^K g_{k}^{(n)}(t)\mathbbm{1}_{\ul{g}_n\in\mathcal{C}_m}}{K \sum_{n=1}^N\mathbbm{1}_{\ul{g}_n\in\mathcal{C}_m}}}$ represents the average satisfaction parameter of Cluster $m$ at time-slot $t$ and $\mathbbm{1}$ being the indicator function.

\label{prop:2}
\end{proposition}
\begin{proof}
See Appendix.
\end{proof}

The above proposition is particularly interesting since it allows clear interpretations to be made. Indeed, for a fixed sequence of prices, it is seen that the best clusters form the famous "Vorono\"{i} cells" in a space which results from an affine transformation of the initial parameter or data space. Through the Kronecker product operation, one can also see that a quantity which matters for obtaining the best clusters is given by the sum of satisfaction parameters, which contrasts with KMC. On the other hand, if the clusters are fixed, the best decisions, which are given by the best representative price profiles have very appealing expressions. \textcolor{black}{The best price is seen to be related to the average satisfaction parameter in an affine manner.} When $a$ is small, the procurement cost becomes almost linear and the best price becomes time-independent and equal to $\displaystyle{\lim_{a \rightarrow 0} x_m^*(t) = b}$. \textcolor{black}{Additionally, when $K \rightarrow \infty$, the optimal price profile is given by $\displaystyle{x_m^*(t) \sim \bar{g}_m(t)}$, which corresponds, at any time, to the spatial average (i.e., over the consumers) of the satisfaction parameters. Therefore, if the provider has access to the spatial average of the satisfaction parameters, it immediately obtains a good approximation of the optimal pricing strategy in the sense of (\ref{eq:power_op2}).} 

%
%
\subsection{Power consumption scheduling}

In this section, the decision-maker is a scheduler. The task of the scheduler is to choose in advance \tc{black}{a sequence of consumption power levels, $\ul{x}= (x(1),...,x(T))$ given some knowledge (e.g., a day-ahead forecast) about the non-controllable part} of the consumption $(g(1),...,g(T))$. The problem of electric vehicle battery charging \cite{Beaude-TSG-2016} given a forecast of the consumption profile associated with the other electric home appliances \textcolor{black}{and the problem of PCS under price uncertainty \cite{Poor-TSG-2011}} typically fall in the setting under consideration. Even in scenarios where the (possibly central) decision entity which computes the consumption profiles, it may be beneficial to cluster the non-controllable profiles $\{\ul{g}_1,\dots,\ul{g}_N\}$ into $M$ groups and find the representative consumption profile for each cluster. This might be typically motivated by complexity issues or for having more robustness regarding the measurement or forecasting noise present in the available non-controllable profiles. A simple but very relevant choice for the performance metric for the scheduler consists in choosing the following function:
\begin{equation}
f_2(\ul{x}; \ul{g})  =    -\|  \mathbf{W} (\ul{x} + \ul{g})    \|_p\
\label{eq:lp_norm_def}
\end{equation}
where $\mathbf{W}$ is a diagonal matrix with non-negative entries and the Lp-norm of a generic vector $v$ of size $T$ is given by $\|v\|_p=(|v_1|^p+\dots+|v_T|^p)^{1/p}$. The matrix is a weighting matrix which may model situations where the price is time-varying. When $p=1$ and $\mathbf{W} = \mathbf{I}_T$ the problem amounts to  minimizing the total energy consumption. When $p=2$, the problem is simple since the electricity price depends on the power consumption in a linear manner. For $p \rightarrow \infty$, minimizing the Lp-norm amounts to minimizing the peak power. Here, we also assume some (classical) constraints on the consumption power: {
\begin{equation}
\begin{split}
&0\leq x(t) \leq x_{\max}\\
&\sum_{t=1}^{T}x(t)\geq E
\end{split}
\end{equation}} where $E >0$ is the energy need. With the notations of Sec. III, this means that the inequality constraint functions write as: $\forall t \in\{1,...,T\}$, $d_{2t-1}(\ul{x}) = -x(t)$, $d_{2t}(\ul{x}) = x(t) - x_{\max}$, and $d_{2T+1}(\ul{x}) = E - \sum_{t=1}^T x(t)$. Note that for $p=1$, the problem is trivial for positive prices and powers. The best decision is obtained by choosing $x(t) = E$ for the time index associated with the lowest coefficient of the diagonal of  $\mathbf{W}$. For $p\geq 2$, the clustering strategy matters and designing a DMOC strategy will be seen to be very beneficial for the performance. The next proposition characterizes the best clusters and representative profiles.



\begin{proposition}[DMOC for PCS] Let $\mc{G} \subseteq \mathbb{R}^{T}$ be the data set. For a given sequence of representative PC profiles $\underline{x}_{m} $ the best way of clustering (in the sense of $f_2$) the \tc{black}{\tc{black}{NCPC}} profile space is given by the following clusters: 
\begin{equation}
\mc{C}_m^*=\left\{ {\underline{g}} \in \mathcal{G}~: \,\|  \mathbf{W}(\ul{x}_m  + \ul{g})   \|_p \leq  \|  \mathbf{W} ( \ul{x}_{m'}  + \ul{g} ) \|_p,  \ \forall m'\neq m\right\}.
\label{eq:clusters-pcs}
\end{equation}
Now, for a given partition of the \tc{black}{\tc{black}{NCPC}} profile space into a set of clusters $\mc{C}_m$, the best representative PC profiles are given by solving the following convex OP:
\begin{equation}
\begin{array}{cl}
\underset{x_m}{\mathrm{minimize}} & \displaystyle{\sum_{n \in  \mc{N}_m}}\| \mathbf{W} ( \ul{x}_m  + \ul{g}_n )  \|_p \\
\mathrm{s.t.} & -x_m(t) \leq 0 \,\,\  \ \ \ \ \ \ \forall t\\ 
                      & x_m(t) - x_{\max} \leq 0 \,\,\forall t\\ 
                      & E - \displaystyle{\sum_{t=1}^{T}} x_{m}(t) \leq 0                 
\end{array}
\label{eq:reps_linfty}
\end{equation}
where $\mathcal{N}_m $, as defined in Sec. III, represent the set of indices of the data which belongs to the cluster $\mathcal{C}_m$.
\end{proposition}
\begin{proof}
See Appendix.
\end{proof}

It is seen that the best clusters (for fixed representatives) have a relatively simple "geometry" since they are generalized Vorono\"{i} cells that is, the Euclidean norm is replaced with the general distance given by the Lp-norm (they coincide for $p=2$). As for the best representatives, here we don't provide an explicit formula. \tc{black}{But the OP to be solved (19) to find them numerically is convex, which strongly facilitates the task of determining them. If complexity to solve this OP or to determine the clusters given by (18) occurred to be an issue, one may resort to approximating the DMOC procedure. \textcolor{black}{Indeed, whatever the actual values for $p$, it is always possible to force $p$ to be equal to $2$ in (18). As a consequence, clusters become Vorono\"i regions. By doing so, one obtains an approximated version of DMOC. The virtue of this approximate version is that it allows one to reduce the complexity as the tesselation/geometry of Vorono\"i regions is known.} The induced performance loss is assessed in the numerical part in a typical scenario (see Fig. 2). }

{\textbf{Remark}: Both RTP and PCS problems are convex. Due to the quadratic structure of the utility functions in RTP, we can provide the expression of the solutions (according to Proposition 4.1.) However, in the PCS problem with Lp norm optimization problem, it is very difficult to express the solution and thus we resort to numerically efficient algorithms (Interior point algorithms) to compute the solution of the optimization problem. In terms of computational complexity, solving the PCS problem is more demanding.}

\section{Numerical performance analysis}
\label{sec:num-perf-analysis}

In the preceding section, several interpretable analytical results have been derived, especially for RTP. To get more insights on the problem of PCS for which less analytical results are available, we dedicate here more space to this case. {All the provided numerical results have been performed by using the \textsf{Matlab} software. In particular, the $k-$means clustering (KMC) technique is executed by using the \textsf{Matlab} routine "kmeans".}


\subsection{Influence of the clustering scheme on the performance of PCS}For all the numerical results concerning the problem of PCS, we consider the peak power minimization problem that is, $p=\infty$ in (\ref{eq:lp_norm_def}). For simplicity reasons, the weighting matrix is chosen as $\mathbf{W}=\mathbf{I}_T$. The database under consideration is the Pecan Street database \cite{pecan}. The used database corresponds to Year 2013 and comprises $N=365$ (non-flexible) household power consumption vectors of size $T=24$ {(with the specific approval by PecanStreet, these consumption profiles are shared in \cite{Data-sharing})}. This database is used to feed Algorithm 1. Algorithm 1 is initialized with randomly chosen representative decision points. The maximum number of iterations is set as $Q=10$. {The  tolerance is set to $T_d=10^{-3}$.} The considered DMOC is given by the set of clusters and representative decisions at convergence. For each household, the DM operation consists in finding a controllable consumption vector $\ul{x}$ minimizing the peak power given a (perfect) day-ahead forecast of the \tc{black}{NCPC} vector $\ul{g}$ (the case of imperfect forecast can be treated by extending our results). Precisely, what is known for taking the decision is to which cluster the \tc{black}{NCPC} vector belongs. The numerical determination of the PC vector is performed by using the dense quasi-Newton Hessian approximation-based interior point technique (implemented by the Matlab \textsc{fmincon} function). The performance of the conventional $k-$means clustering technique and the proposed DMOC technique are measured in terms of the function $f_2$ (see (\ref{eq:lp_norm_def})); more precisely, unless stated otherwise, the latter is averaged over several randomly selected household profiles namely, Households 4998, 6910, 9499, and 9609. The energy need in terms of PC for a household is set to $E=30$ kWh. 
     
\textcolor{black}{First, we want to assess the loss induced by clustering (namely, by using a fixed number of possible decisions $\ul{x}$ instead of using the optimal solution $\ul{x}^*(\ul{g})$ for every $\ul{g}$)}. For this, we define the \textit{relative optimality loss} of the generic clustering scheme $\mathrm{C}$ with respect to the ideal case as follows ${\rho_{2,\mathrm{C}} (\%) = \frac{F_2^{\text{perfect}} - F_2^{\mathrm{C}}}{F_2^{\text{perfect}}}\times 100}$ where $F_2$ is obtained by averaging $f_2$ over several realizations of the \tc{black}{NCPC} vector and the performance of the ideal case is attained \textcolor{black}{by assuming that the optimal PC vector $\ul{x}^*$ is available (this amounts to having an infinite number of clusters)}. \tc{black}{The  natural relevance of the notion of relative optimality loss stems from the fact the decision-making entity process is represented by the maximization of the function $f_2$. Therefore, what matters is that the decision taken is as close as possible to the ideal situation which is obtained by maximizing $f_2(\ul{x};\ul{g})$ with an absolutely perfect knowledge of the parameters $\ul{g}$. Here, $F_2$ intervenes instead of $f_2$ because the performance is averaged.} Fig.~3 represents $\rho_{2,\mathrm{C}} (\%)$ as a function of the number of clusters for 4 different clustering schemes when $F_2$ is obtained by averaging over the 365 daily \tc{black}{\tc{black}{NCPC}} profiles of Household 9499. Indeed, DMOC is compared to four popular clustering schemes namely, KMC, hierarchical clustering (HC), fuzzy C-means clustering (FCMC), and {symbolic aggregate approximation (SAX) based clustering \cite{Lin-2003}\cite{Fonseca-2016}}. For HC, the squared Euclidean distance and weighted pair group method with arithmetic mean are used. For FCMC, the fuzzifier exponent parameter is set to 2. {For SAX based clustering, the window size is set to $4$ and the alphabet size is fixed to $8$.} Regarding the performance, for $15-20$ clusters, the optimality loss for KMC, HC, FCMC and SAX based clustering are seen to be around $20\%$. With the proposed approach (DMOC), it is seen that the optimality loss is as low as $2-3\%$ for the same number of clusters, which represents a very significant improvement. \tc{black}{Additionally, by approximating the clusters by Vorono\"i regions, the approximated DMOC allows one to reduce  complexity  while only inducing a reasonable performance loss w.r.t. the original DMOC.} To better illustrate the other potential benefits from using DMOC, we mainly show the comparison between DMOC and KMC in the following figures. In Fig.~4, the desired maximum peak power level is fixed to a given value in the range $[3.8, 4.4]$ kW. Then, one computes the number of clusters which allows one to guarantee that the total power will not exceed this value. Fig.~4 shows, in particular, that the required number of clusters can be very high when the constraint on the maximum power level is strong (e.g., when it equals $3.8$ kW). On the other hand, using DMOC allows one to adapt in an ideal manner the shape of the clusters and the representative decisions, which explains why the number of required clusters can be made very small. To better understand how DMOC operates in terms of shaping the clusters and selecting the representative decisions, we consider in the next subsection special cases allowing to make interpretations.

\begin{figure}[h]
   \begin{center}
        \includegraphics[width=.5\textwidth]{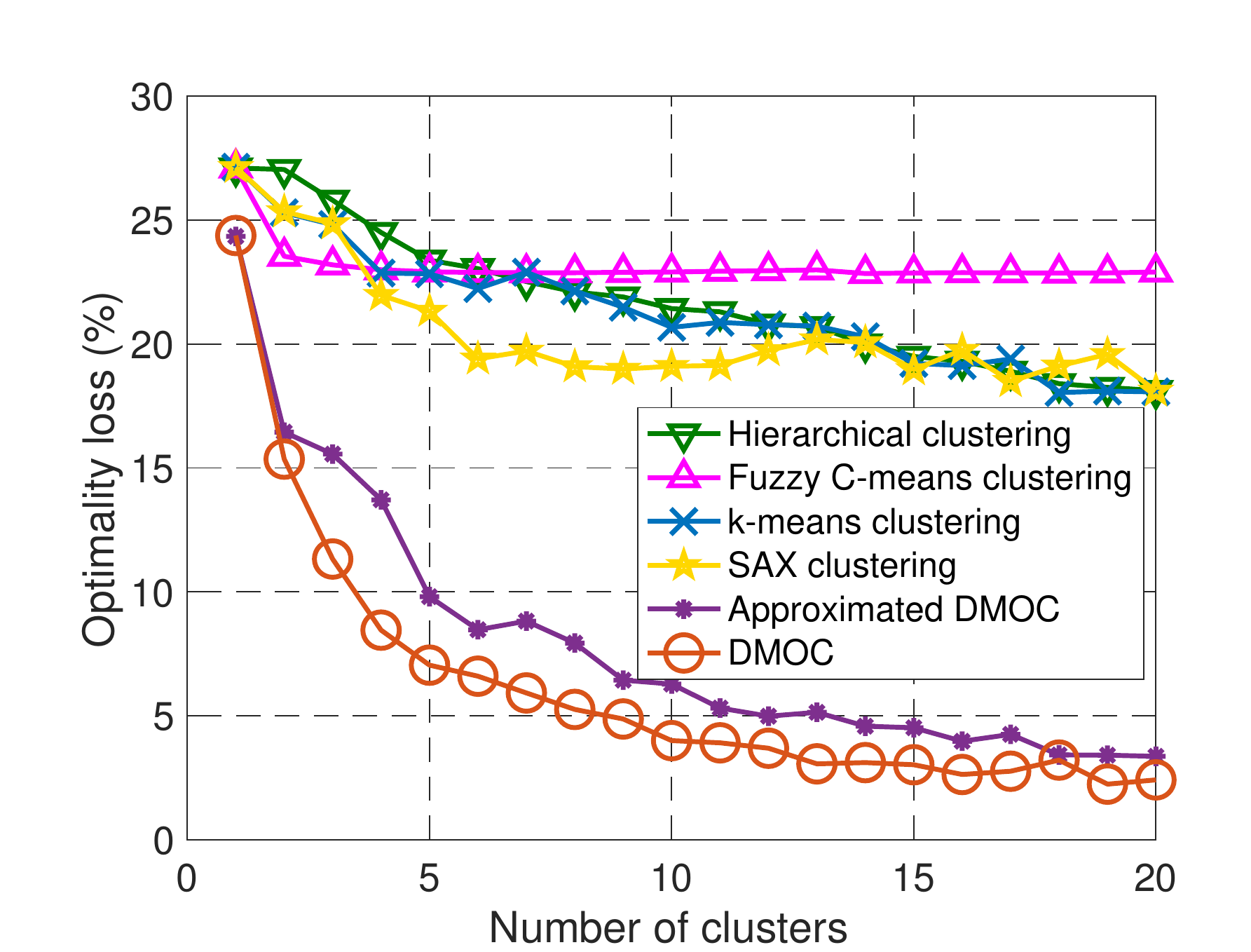}
    \end{center}
    \caption{Relative optimality loss induced by the finiteness of the number of clusters.}
\end{figure}

\begin{figure}[h]
   \begin{center}
        \includegraphics[width=0.5\textwidth]{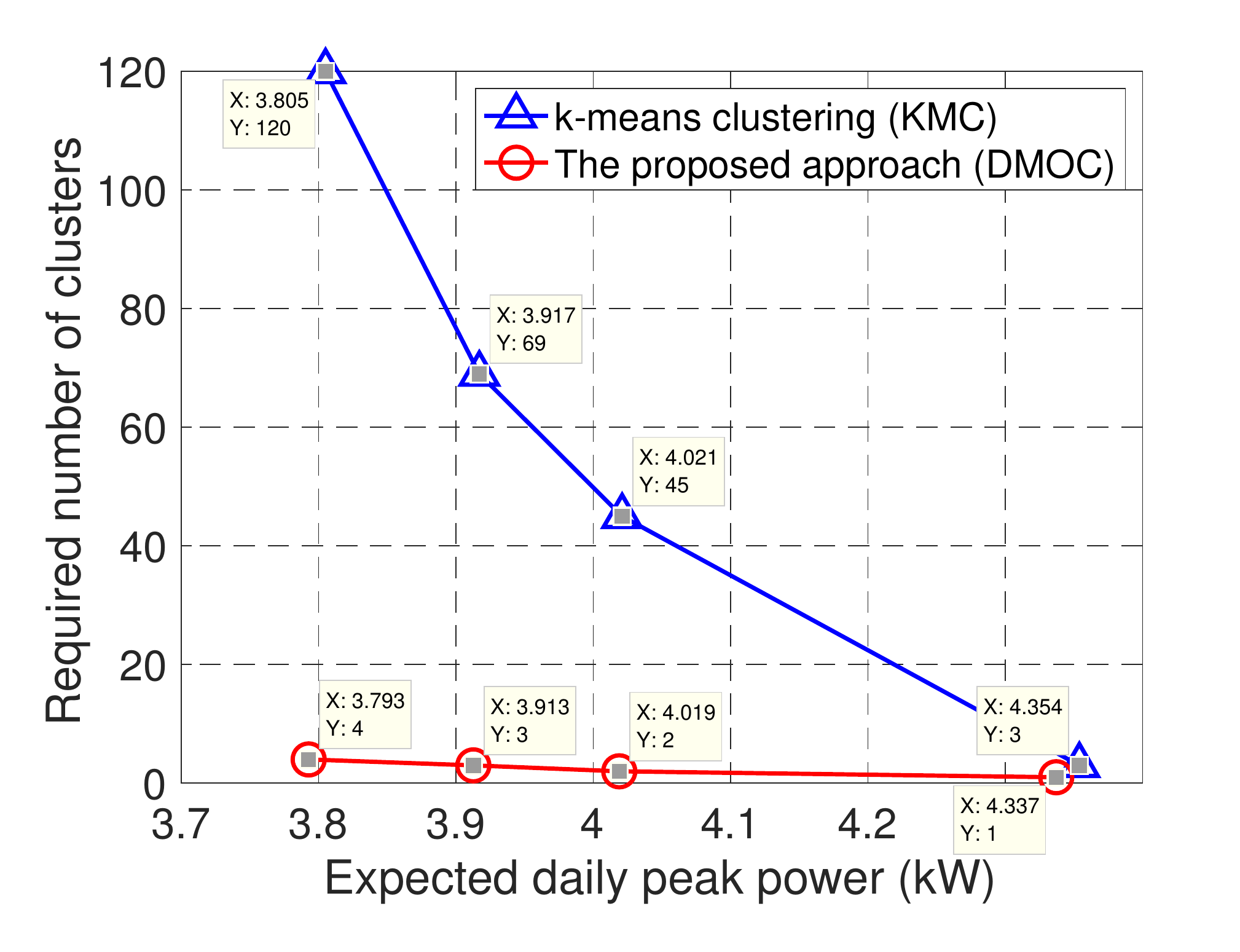}
    \end{center}
        \caption{Required number of clusters to reach a given target in terms of maximum peak power.}
\end{figure}

\subsection{About the shape of DMOC clusters and DMOC representative decisions (PCS)}

To be able to represent the clusters geometrically, we fix the dimension of the data space to $T=2$. This means that the consumption profile $\ul{g}$ comprises 2 phases with constant power; here, it corresponds to the average power from 6 am to 6 pm and that from 6 pm to 6 am, always for the Pecan street database. As a consequence, the number of clusters is also small. It is set as $M=4$. For this setting, Fig.~5 shows the clusters obtained when using \textcolor{black}{DMOC (left subfigure) and KMC (right subfigure)}. With KMC, the obtained clusters correspond to Vorono\"{i} cells. With DMOC, the obtained clusters are markedly different. The latter are tailored to the $L_{\infty}-$norm. These clusters are much more suited to manage the peak power reduction problem. Roughly, data samples are grouped into regions in which the difference in terms of power between the two consumption phases (namely, the quantity ($|g_n(1) - g_n(2)|$) is small. Now, let us turn our attention to the shape of representative PC profiles. In this case, any value for the data space dimension can be assumed. {Therefore, we assume the typical value $T=24$ and that the data are clustered in three groups that is, $M=3$. The rationale behind this choice is to make apparent the main features of interest that are automatically extracted by the DMOC technique. As far as the decision to be taken aims at minimizing the peak power, the main feature is found to be the time information associated with the occurrence of the most likely dominant peak power.} Fig.~6 depicts the three PC profiles (in red dash line) of KMC and DMOC, respectively. \tc{black}{As a side information, for each of the clustering approaches (KMC/DMOC), the empirical mean of the \tc{black}{NCPC} profiles (in solid blue line) over the cluster under consideration is given for each cluster. Notice that, since the clusters provided by the two approaches differ, the means also differ.} \textcolor{black}{It can be seen that DMOC classifies the \tc{black}{\tc{black}{NCPC}} $\ul{g}$ according to the peak power occurrence time. The peak of the first \tc{black}{\tc{black}{NCPC}} profile (called Type I) occurs in the afternoon while the peaks of the second and the third type occurs in the early evening and late evening, respectively. The representative PC profiles naturally comprise higher values over off-peak periods of its corresponding \tc{black}{\tc{black}{NCPC}} profiles. By contrast, KMC provides less suited PC profiles by considering the $L_2-$norm of the \tc{black}{\tc{black}{NCPC}} profiles instead of adapting to the decision performance metric, here an $L_p-$norm.}  {Fig.~\ref{fig:comparison_member_vs_representative} allows one to be able to compare the representative profile of a cluster with the rest of the cluster members. Interestingly, the time-slot of the representative peak (right figures) corresponds to the time-slot which has the highest probability of peak power occurrence (the densest part in the left figures).}


\begin{figure}[h]
\hspace{-6mm}
	\centering
	\begin{subfigure}[t]{.50 \linewidth}
		\includegraphics[width=1.15\linewidth, height=1.1\linewidth]{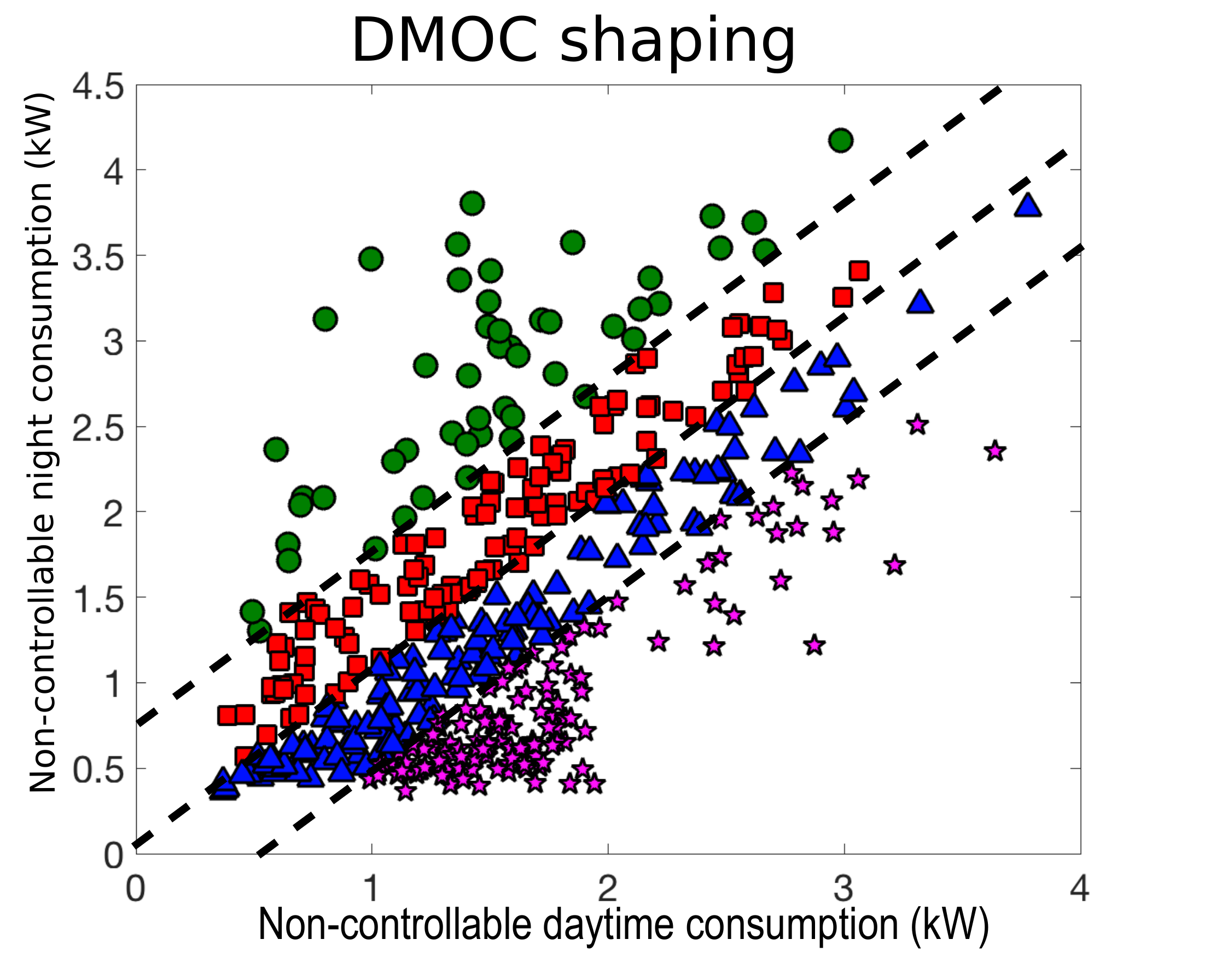}
	\end{subfigure}\,\,
	\begin{subfigure}[t]{.50 \linewidth}
		\centering
		\includegraphics[width=1.15 \linewidth, height=1.1\linewidth]{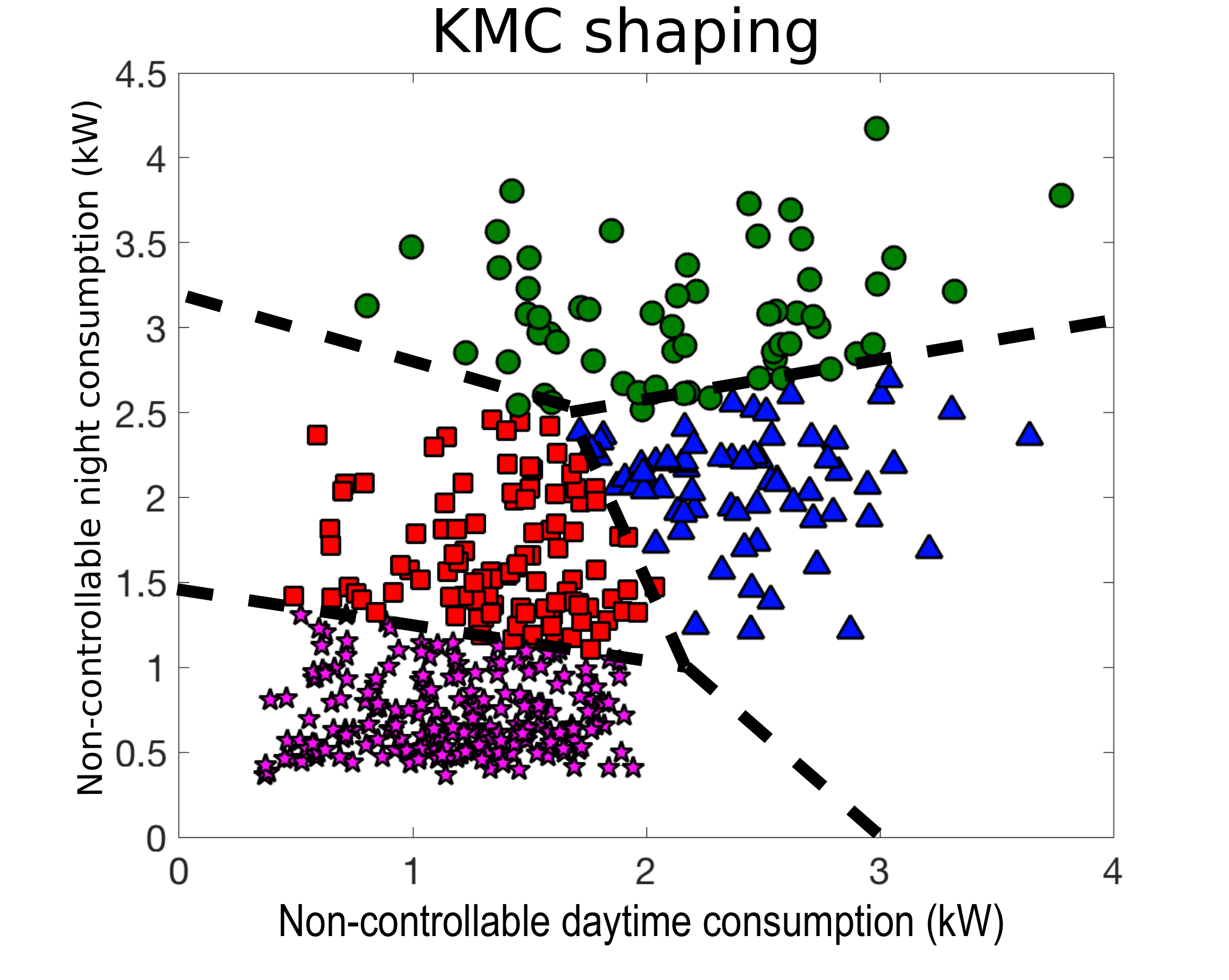}
	\end{subfigure}
	    \caption{Geometry of the cluster shape.}
\end{figure}

\begin{figure}
   \centering
\begin{tabular}{lccc}
&\includegraphics[width=.23\textwidth]{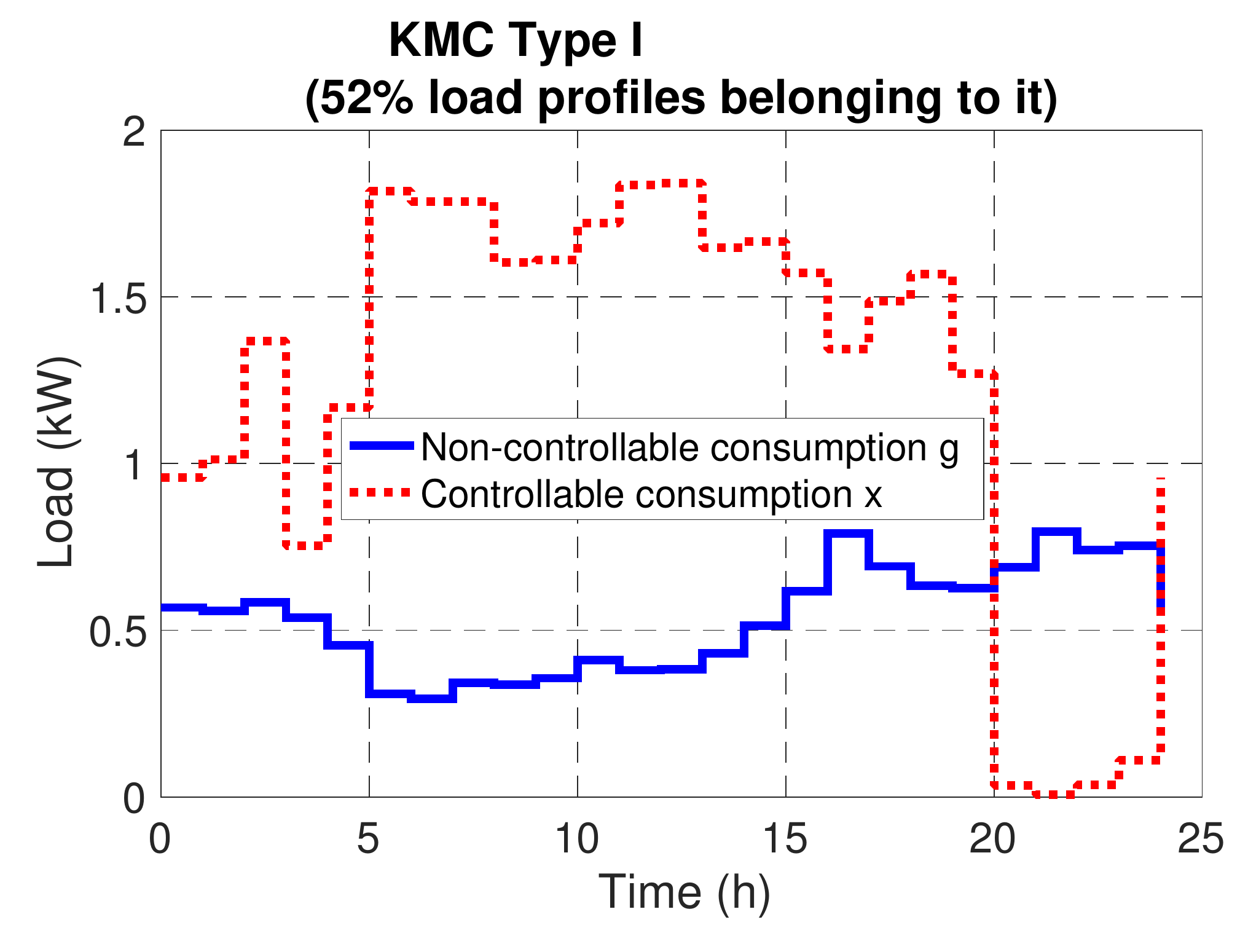}&\includegraphics[width=.23\textwidth]{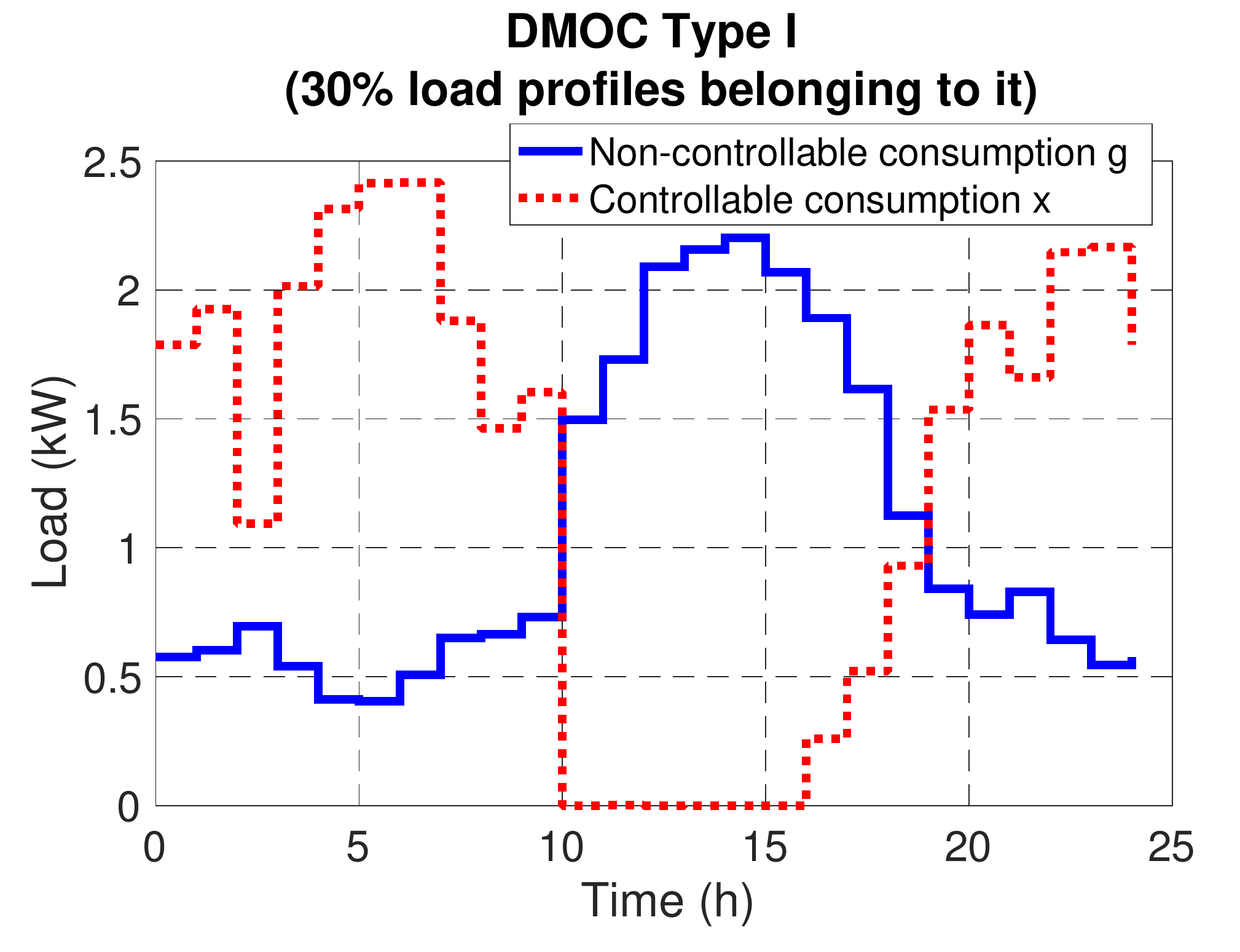}\\&\includegraphics[width=.23\textwidth]{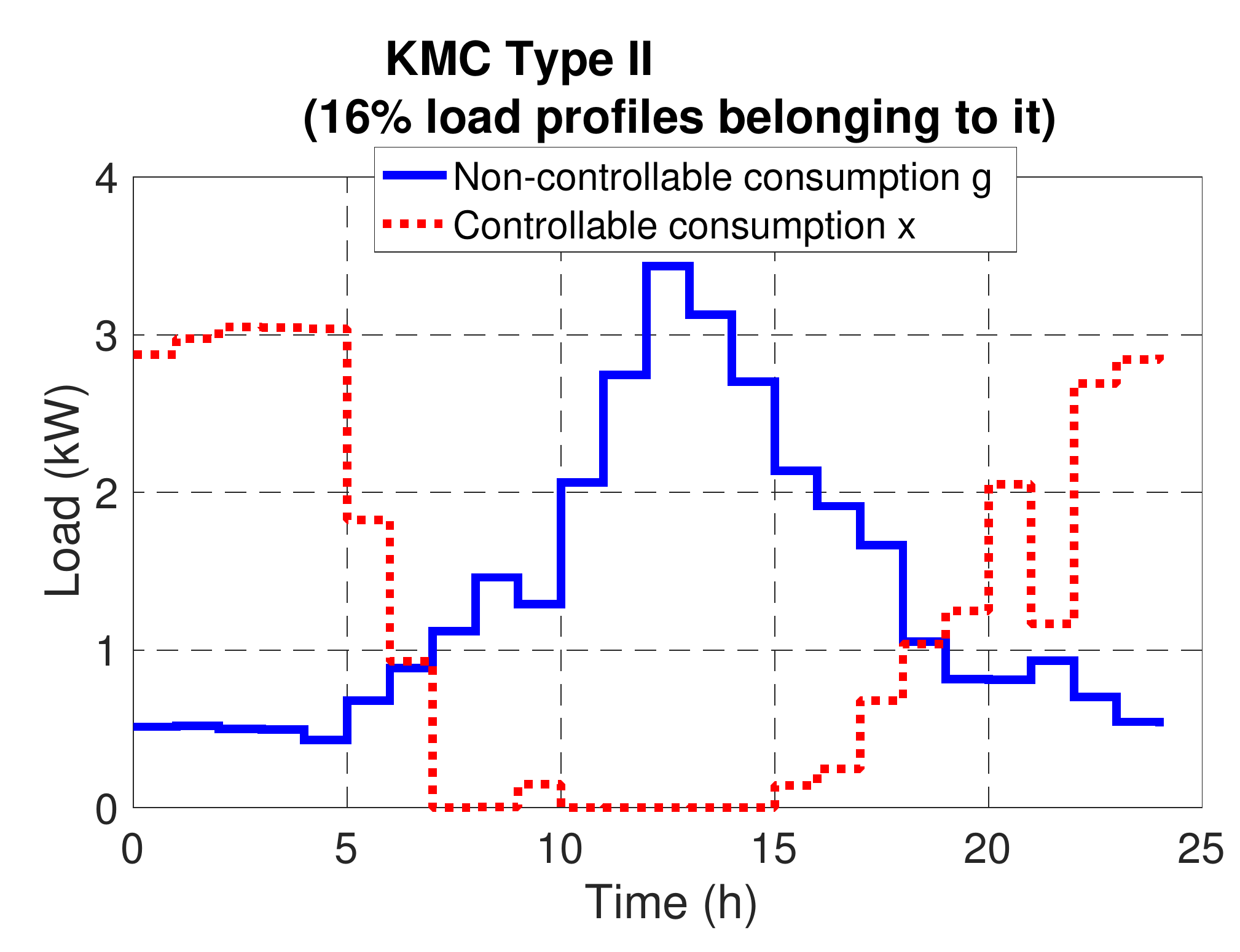}&\includegraphics[width=.23\textwidth]{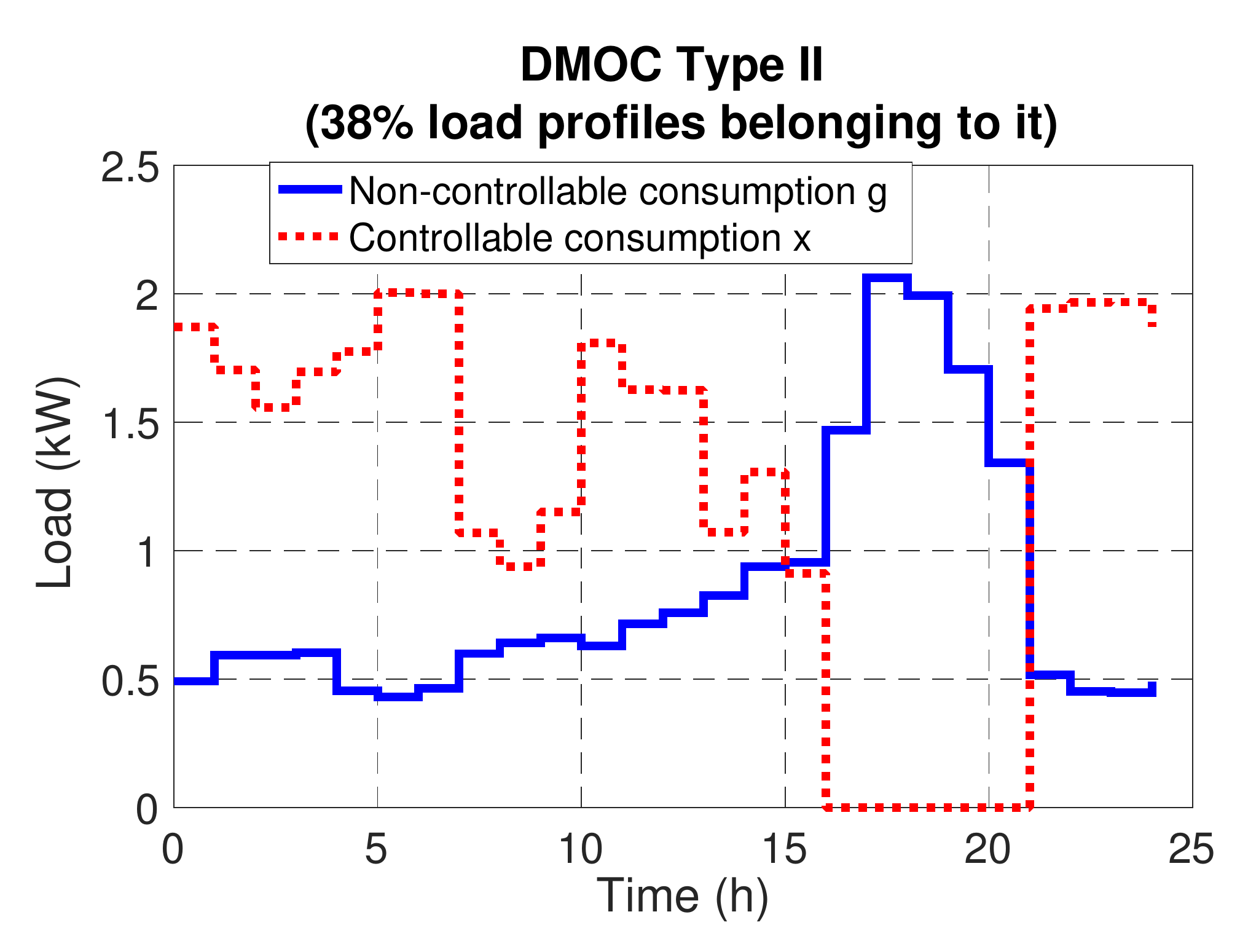}\\&\includegraphics[width=.23\textwidth]{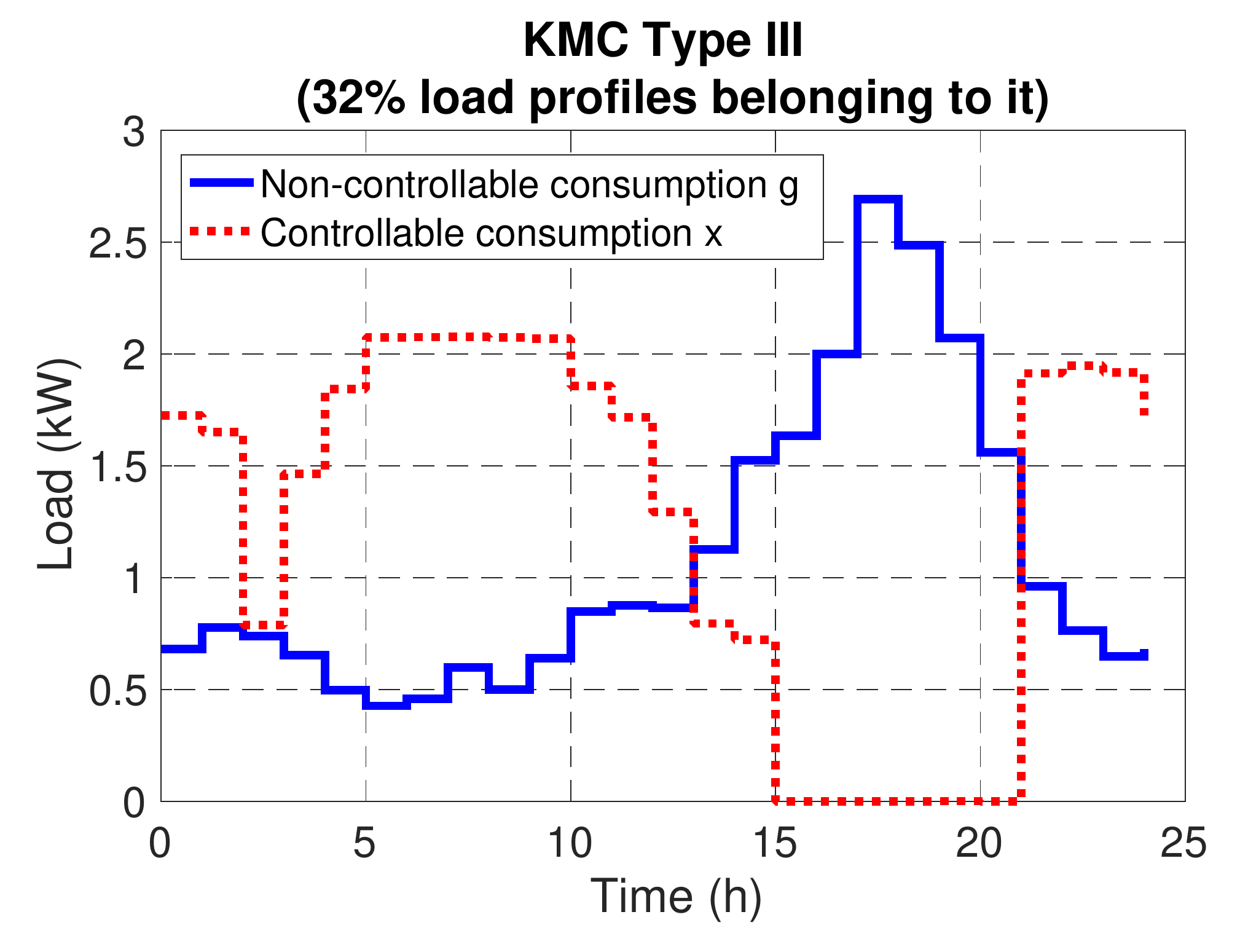}&\includegraphics[width=.23\textwidth]{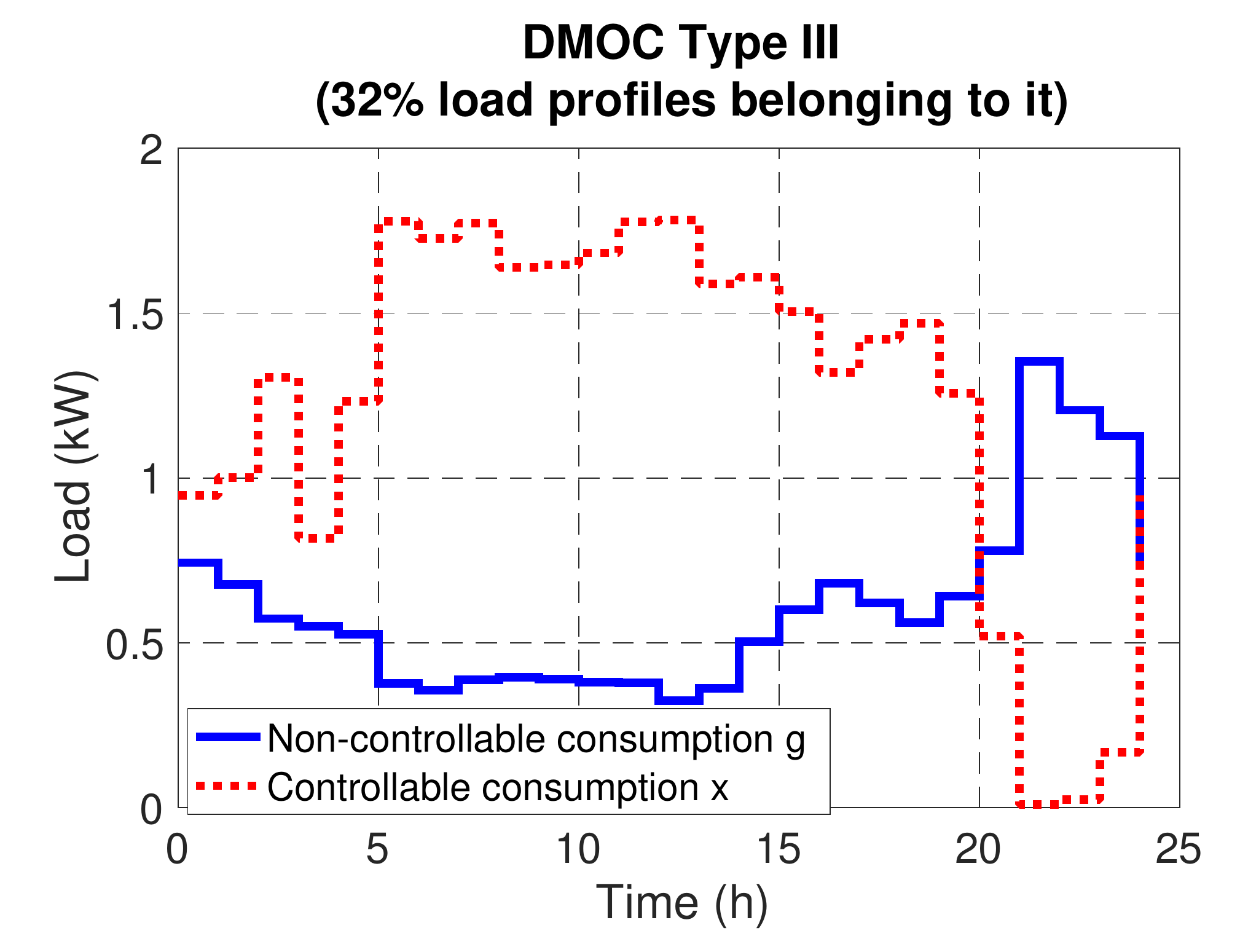}
\\
\end{tabular}

    \caption{DMOC vs KMC. Three PC profiles and their corresponding average \tc{black}{\tc{black}{NCPC}} profiles are represented.}
    \label{fig:fig1} 
\end{figure}

\begin{figure}[h]
   \begin{center}
        \includegraphics[width=.48\textwidth]{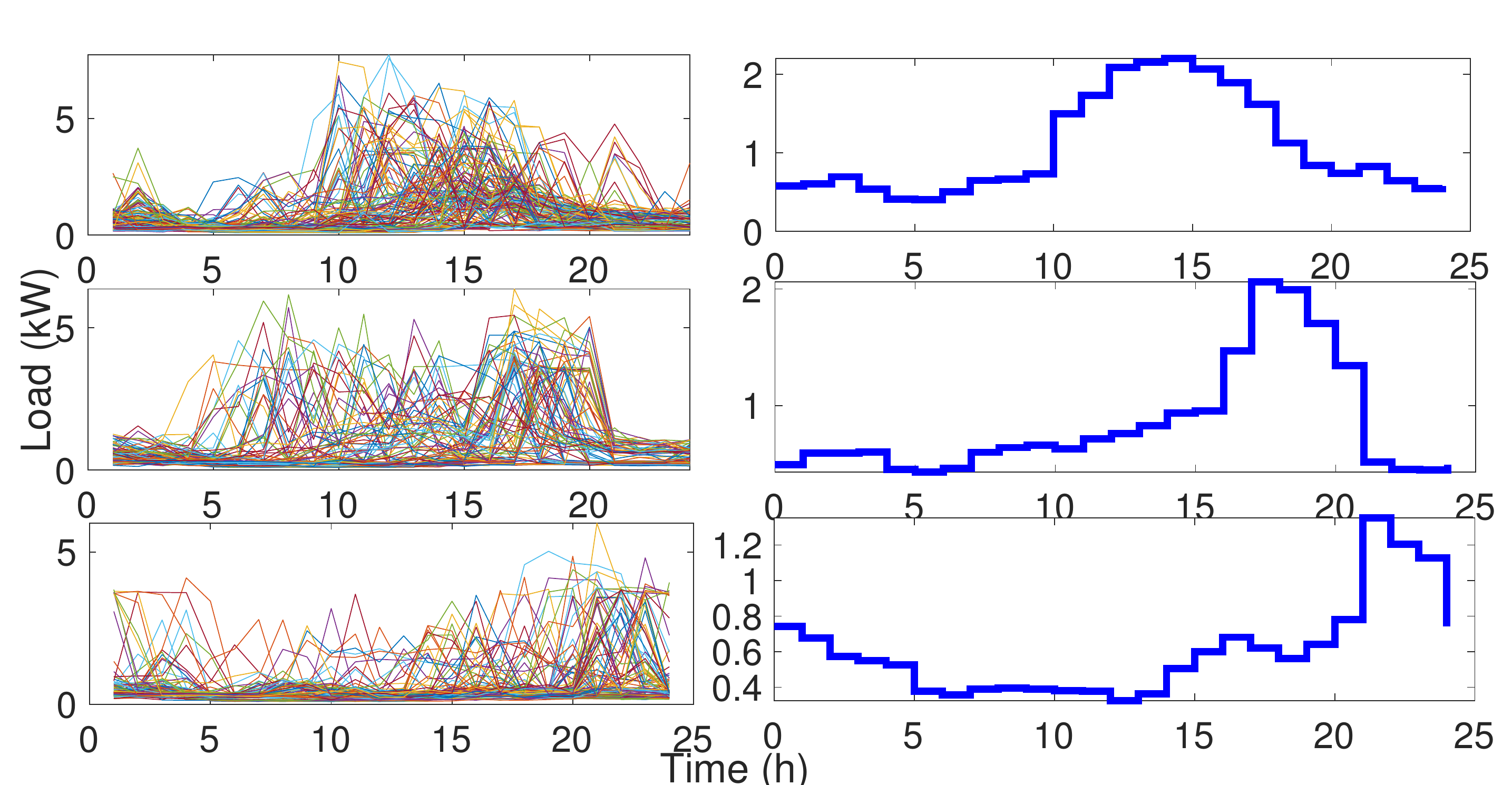}
    \end{center}
    \caption{Comparison between cluster representative and cluster members}
    \label{fig:comparison_member_vs_representative}
\end{figure}

\subsection{Influence of the data on the performance gains (PCS)}

In the previous subsections, results were averaged over four randomly selected households. Here, we look at the performance for each household, in particular, our goal being to see to what extent the nature of the data influences the outcome in terms of gains brought by DMOC over KMC. Fig.~8 represents the relative optimality loss $\rho_{2,\mathrm{C}}$ for \textcolor{black}{given realizations of the \tc{black}{NCPC} profile with different households}. The four curves correspond to the randomly selected households. It is seen that the loss induced by clustering (here only DMOC is considered) clearly depends on the household but is always as low as $5\%$ when the number of clusters exceeds $10$. Interestingly, we have seen that the entropy of a non-flexible consumption profile can be used as a measure to know whether DMOC will bring a significant performance gain. Indeed, by denoting $\widehat{p}(t)$ the empirical probability that the non-flexible peak power occurs at time-slot $t$ by $\displaystyle{\widehat{p}(t)=\frac{1}{N} \sum_{n=1}^N \mathbbm{1}_{g_n(t)=\max \ul{g}_n}}$, the \textit{entropy} of an \tc{black}{NCPC} profile merely expresses as 
\begin{equation}
\displaystyle{H(\widehat{p})=- \sum_{t=1}^{T}\widehat{p}(t)\log_2 \widehat{p}(t)}.
\end{equation} For Households 6910, 4998, 9609, and 9499, the value of the entropy is respectively given by 3.45, 3.82, 3.91, and 4.19. This shows here that entropy may reflect well the optimality loss obtained when using DMOC. 

\begin{figure}[h!]
   \begin{center}
        \includegraphics[width=.5\textwidth]{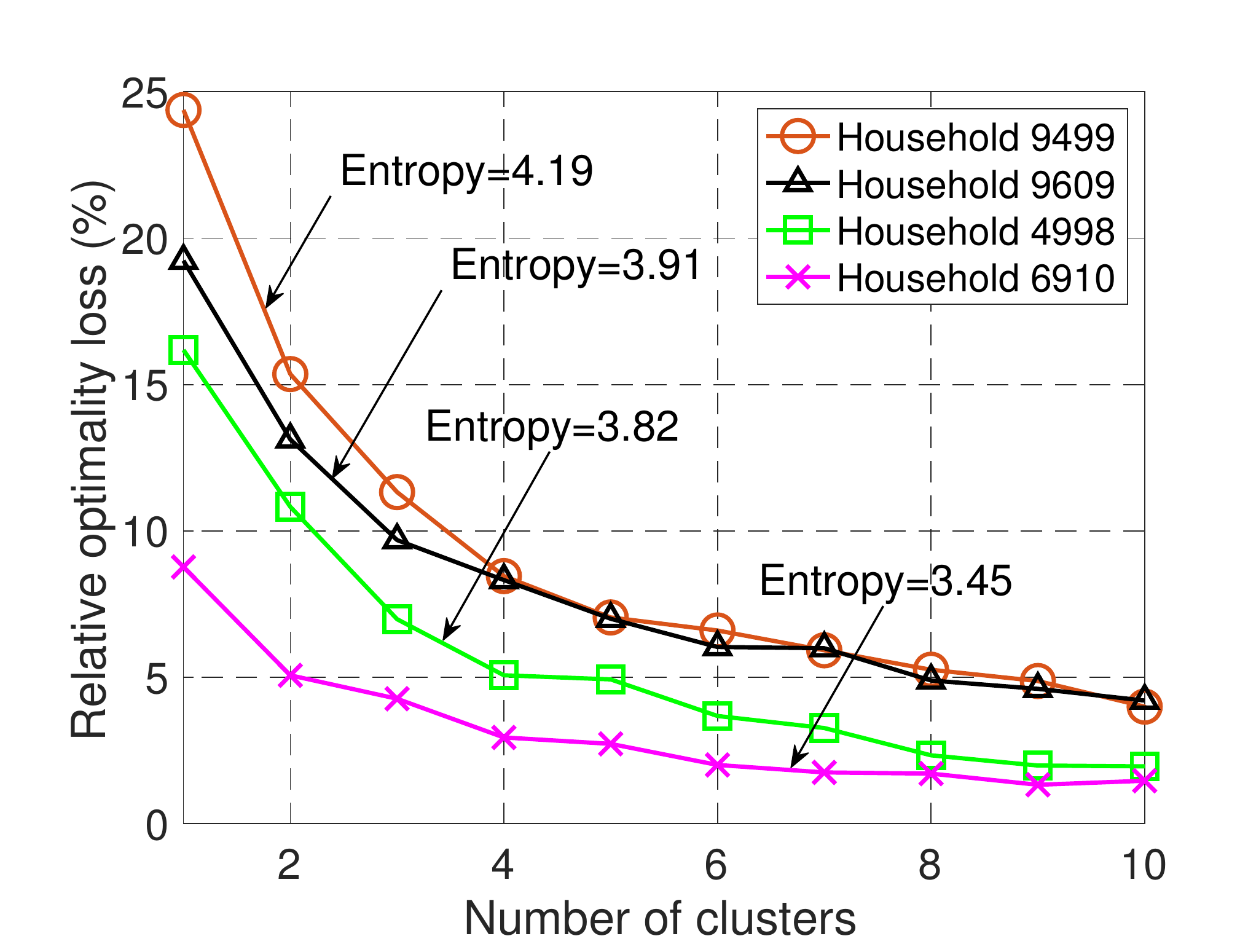}
    \end{center}
    \caption{Relative optimality loss against the number of clusters.}
        \label{fig:comparison_household}
\end{figure}

\subsection{Potential benefits from using DMOC for RTP}

Here, we consider the problem of RTP. The simulation setting we choose is very close to \cite{MR-SGC-2010}. We consider a system with a unique provider and $K=5$ or $K=10$ consumers. For each day, the consumer satisfaction parameter $g_k(t)$ is assumed to be constant for a period of $6$ hours, which means that $\ul{g}_k = (g_k(1),...,g_k(T))$ with $T=4$. Additionally, the satisfaction parameters $g_k(t)$ are assumed to be realizations that are i.i.d. over the consumers and time. Each $g_k$ is uniformly distributed over the interval $[2,3]$ and we choose $\alpha=0.5$, $a=0.1$, $b=0$, and $c=10$ for the parameters of the function $f_1$ (see (\ref{eq:welfare})). Fig.~9 represents the relative optimality loss $\rho_{1,\mathrm{C}} (\%) = \frac{F_1^{\text{perfect}} - F_1^{\mathrm{C}}}{F_2^{\text{perfect}}}\times 100, \quad \mathrm{C}\in\ \{\mathrm{KMC},\mathrm{DMOC}\}$ as a function of the number of clusters for KMC and DMOC where $F_1$ corresponds to an average of $f_1$ over $N=365$ draws for the vector of satisfaction parameters. The gain brought by DMOC over KMC is globally less significant than for the peak power minimization problem; this can be explained by the quadratic structure of the problem, which implies that the DMOC representatives are also obtained by using the Euclidean distance just as KMC does. \textcolor{black}{However, as it can be seen from (\ref{eq:best_representative_case1}), exploiting an affine transformation of the initial parameters, DMOC considers the sum of all the consumers' demand levels as a single parameter. By classifying the dataset according to this automatically extracted feature, the optimality loss can be made significantly lower compared to the conventional approach such as KMC. If one wants to guarantee small optimality losses (say $< 10\%$) it is even seen that it may be impossible for KMC to reach the corresponding accuracy level, making the use of DMOC necessary.}
\begin{figure}[h]
   \begin{center}
        \includegraphics[width=.5\textwidth]{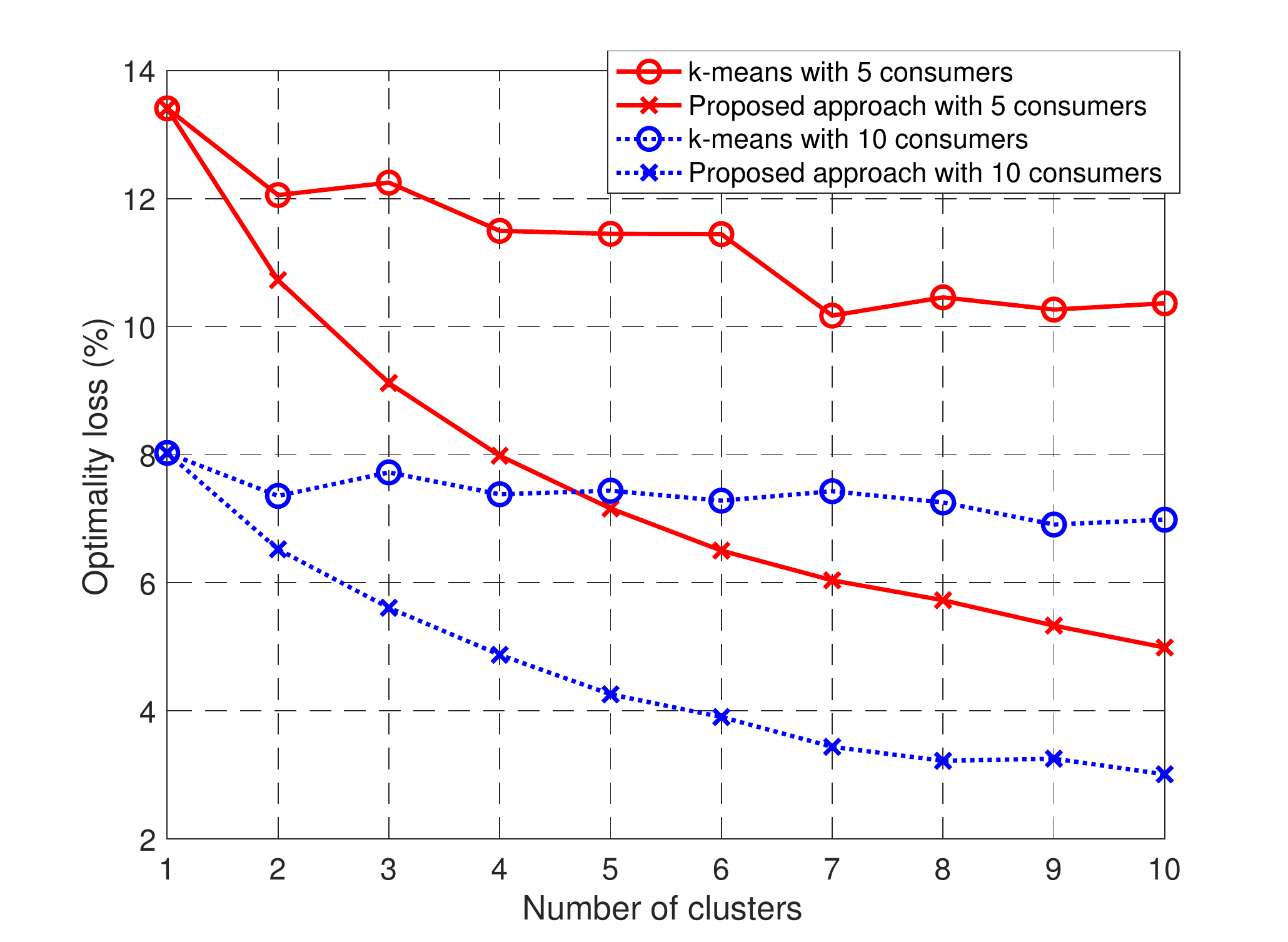}    \end{center}
    \caption{\textcolor{black}{Relative optimality loss against the number of clusters for the problem of RTP.}}
\end{figure}
\section{Conclusion}
In this paper, we have provided a new approach to clustering that allows one to extract ex ante and in an automatic manner, for any performance metric, the features relevant to the decision maker using the data. By doing so, one can minimize the impact of the finiteness of the number of clusters on the final performance. For instance, for the peak power problem, we have seen that the number of required clusters to perform the corresponding power scheduling task can be divided by a factor as high as 30 compared to conventional clustering. The analytical results provided for the problem of \tc{black}{real-time pricing} and \tc{black}{power consumption scheduling} illustrate very well the effects of the adopted point of view compared to the conventional point of view. The numerical analysis clearly illustrates the benefits of decision-making oriented clustering e.g., in terms of required number of clusters or optimality loss for the decision making process. The proposed approach might be refined. For instance, an interesting and deepened discussion of the complexity issue might be conducted. For a given complexity level for the clustering plus decision operation, the conventional approach might be compared to the proposed approach. For this, approximation-based low-complexity \tc{black}{decision-making oriented clustering} schemes may be considered. Also, the impact of the choice of the performance metric on the performance gain of \tc{black}{decision-making oriented clustering} should be investigated more; we provide the answer for two famous performance metrics but a deeper problem would be to mathematically characterize functions for which the gain is large, intermediate, or small. The extension to the case where the data used by the decision-maker are noisy would be very relevant; possible paths would be to generalize the proposed algorithm the way the Lloyd-Max quantization algorithm has been generalized to noisy inputs  or to exploit reinforcement learning algorithms with noisy measurements.

\section{Appendix*}

\subsection{Proof of Prop. 3.1}

\begin{proof}
The OP associated with (\ref{sec:new-Q}) can be rewritten as:
\begin{equation}
\begin{split}
\underset{  \left( \mathcal{N}_1,..., \mathcal{N}_M, \ul{r}_1,..., \ul{r}_M  \right)  }{\mathrm{minimize}}&-\quad\sum_{m=1}^M\sum_{n\in \mathcal{N}_m} f(\ul{x}^{\star}(\ul{r}_m);\ul{g}_n)\\
\mathrm{s.t.}&\quad d_i(\ul{x}^{\star}(\ul{r}_m))\leq 0\\
&\quad e_j(\ul{x}^{\star}(\ul{r}_m))=0
\end{split}.
\end{equation}
By replacing $\ul{x}^{\star}(\ul{r}_m)$ with $\ul{x}_m$, the equivalence is proved.
\end{proof}

\subsection{Proof of Prop. 4.1}

\begin{proof}
By plugging $\ell^{\star}_k(x(t))=\frac{g_k(t) -x(t) }{\alpha}$ into (\ref{eq:welfare}), the  function $f_1(\ul{x};\ul{g})$ can be rewritten as:{\footnotesize \begin{equation}
\label{eq:proof_prop1}
\begin{split}
&f_{1}(\ul{x}; \ul{g}) \\
=& \sum_{t=1}^{T} \left[ -\tilde{a}(x(t)-\frac{1}{2\tilde{a}\alpha^2}(\alpha bK+2aK\sum_{k=1}^Kg_k(t)))^2+c_t(\ul{g})\right]\\
=&-\tilde{a}\| \mathbf{A} \underline{g}  + \textcolor{black}{\underline{\beta}}    - \underline{x}  \|_2^2+ \sum_{t=1}^{T}c_t(\ul{g})
\end{split}
\end{equation}}
where \textcolor{black}{$c_t(\ul{g})=\sum_{k=1}^K\left[\frac{(\alpha-2a)g_k^2(t)}{2\alpha^2}-\frac{bg_k(t)}{\alpha}\right]$} is independent of $\ul{x}$ and thus irrelevant for the choice of $\mathcal{C}_m$.
By combining (\ref{eq:proof_prop1}) and (\ref{eq:vq2_vector}), for given representatives, the optimum regions can be written as (\ref{eq:cell_caseII}).

For given clusters ($\mathcal{C}_1$,\dots,$\mathcal{C}_M$), the best representative is obtained by solving the following OP:
\begin{equation}
\ul{x}_m^*\in\quad -\underset{\ul{x}\in\mathcal{X}}{\arg\min}\displaystyle\sum_{n\in {\mathcal{N}_m}} f(\ul{x}_m;\ul{g}_n).
\end{equation}
In the RTP case, the sum-utility expresses as:
{\scriptsize \begin{equation}
\begin{split}
&\sum_{n\in {\mathcal{N}_m}}f_{1}(\ul{x}_m; \ul{g}_n) \\
=& \sum_{t=1}^{T} \sum_{n\in {\mathcal{N}_m}}\left[ -\tilde{a}(x_m(t)-\frac{1}{2\tilde{a}\alpha^2}(\alpha bK+2aK\sum_{k=1}^Kg_{k}^{(n)}(t)))^2+c_t(\ul{g}_n)\right]\\
=&\sum_{t=1}^{T} \sum_{n=1}^N\mathbbm{1}_{\ul{g}_n\in\mathcal{C}_m}\left[ -\tilde{a}(x_m(t)-\frac{1}{2\tilde{a}\alpha^2}(\alpha bK+2aK\sum_{k=1}^Kg_{k}^{(n)}(t)))^2+c_t(\ul{g}_n)\right]\\
=&\sum_{t=1}^{T} \left[ -\tilde{a}(\sum_{n=1}^N\mathbbm{1}_{\ul{g}_n\in\mathcal{C}_m})\left((x_m(t)-\frac{a \ol{g}_m(t) + \frac{\alpha b}{2K} }{a + \frac{\alpha}{2K}})^2+c^{\prime}_t(\ul{g}_n)\right)\right].
\end{split}
\end{equation}}
Therefore, maximizing $\sum_{n\in {\mathcal{N}_m}}f_{1}(\ul{x}_m; \ul{g}_n)$ is equivalent to minimizing $(x_m(t)-\frac{a \ol{g}_m(t) + \frac{\alpha b}{2K} }{a + \frac{\alpha}{2K}})^2$,  the solution  can be obtained and written as (\ref{eq:best_representative_case1}).
\end{proof}

\subsection{Proof of Prop. 4.2}

\begin{proof}
The result follows from replacing $f$ with the $L_p-$norm function $f_2$ and by noticing that the problem is convex since $\|\ul{x}\|_p$ is convex w.r.t. $\ul{x}$.
\end{proof}

\tc{black}{\section*{Acknowledgement}
This work is partially supported by the RTE-CentraleSupelec Chair.}

\end{document}